\pdfoutput=1
\documentclass[11pt]{article}

\usepackage{acl}

\usepackage{times}
\usepackage{latexsym}
\usepackage{color}
\usepackage{bm}
\usepackage{amsmath}
\usepackage{enumitem}
\usepackage{amssymb}
\usepackage{setspace}
\usepackage{multirow}
\usepackage{pifont}
\usepackage{tikz}
\usepackage{xcolor}
\usepackage{subcaption}
\usepackage{tikz}
\definecolor{tan}{rgb}{0.937, 0.902, 0.843}
\usepackage{colortbl} % For row colors
\usepackage{pifont}
\usepackage{bm}
\usepackage{amssymb}
\usepackage{amsmath}
\usepackage{amssymb}
\usepackage{amsmath}
\usepackage[noend]{algorithmic}
\usepackage[linesnumbered,ruled,vlined]{algorithm2e}
\usepackage[switch]{lineno}
\usepackage{xcolor}
\usepackage{color}
\usepackage{bbding}
\usepackage{graphicx}
\usepackage{booktabs}
\usepackage{caption}
\usepackage{subcaption}
\usepackage{multirow}
\usepackage{subcaption}
\usepackage{tikz}
\definecolor{tan}{rgb}{0.937, 0.902, 0.843}
\usepackage{colortbl} % For row colors
\usepackage{pifont}
\usepackage{amsmath,amssymb,bm} % 加载 bm 宏包用于加粗数学符号
\usepackage[most]{tcolorbox}

\usepackage{threeparttable}
\usepackage{graphicx}
\usepackage{amsmath}
\usepackage{amssymb}
\usepackage{underoverlap}
\usepackage[T1]{fontenc}
\usepackage[utf8]{inputenc}

\usepackage{microtype}

\title{Bridging Social Psychology and LLM Reasoning: Conflict-Aware Meta-Review Generation via Cognitive Alignment}

\author{
	Wei Chen$^{1*}$, Han Ding$^{1}$\thanks{~~Equal Contributions.}, Meng Yuan$^{1}$, Zhao Zhang$^{2}$, Deqing Wang$^{1}$, Fuzhen Zhuang$^{1,3}$\thanks{~~Corresponding Author.} \\
	$^{1}$Beihang University, China \\
	$^{2}$Chinese Academy of Sciences,  China \\
	$^{3}$Zhongguancun Laboratory, China \\
	\{chenwei23, zhuangfuzhen\}@buaa.edu.cn \\
}

\begin{document}
		\maketitle
		\begin{abstract}
		The rapid growth of scholarly submissions has overwhelmed traditional peer review systems, driving the need for intelligent automation to preserve scientific rigor. 
		While large language models (LLMs) show promise in automating manuscript critiques, their ability to synthesize high-stakes meta-reviews, which require conflict-aware reasoning and consensus derivation, remains underdeveloped. Existing methods fail to effectively handle conflicting viewpoints within differing opinions, and often introduce additional cognitive biases, such as anchoring effects and conformity bias.
		To overcome these limitations, we propose the Cognitive Alignment Framework (CAF), a dual-process architecture that transforms LLMs into adaptive scientific arbitrators. By operationalizing Kahneman's dual-process theory, CAF introduces a three-step cognitive pipeline: review initialization, incremental integration, and cognitive alignment.
		Empirical validation shows that CAF outperforms existing LLM-based methods, with sentiment consistency gains reaching up to 19.47\% and content consistency improving by as much as 12.95\%.
	\end{abstract}
	
	\section{Introduction}
	Academic peer review is fundamental to ensuring scientific rigor, with experts assessing research validity through structured evaluation protocols~\cite{chubin1990peerless,rowland2002peer,alberts2008reviewing}. However, the rapid rise in scholarly submissions has created challenges in maintaining these standards, leading to imbalances between reviewer workload and quality assurance~\cite{lee2013bias,fox2017recruitment}. This scalability issue has driven a shift towards intelligent automation, with large language models (LLMs) showing great promise in enhancing peer review through multimodal reasoning and domain-specific text generation~\cite{tyser2024ai,jin2024agentreview,zhou2024llm}.
\begin{figure}[t]
	\centering
	\setlength{\fboxrule}{0.pt}
	\setlength{\fboxsep}{0.pt}
	\fbox{
		\includegraphics[width=1\linewidth]{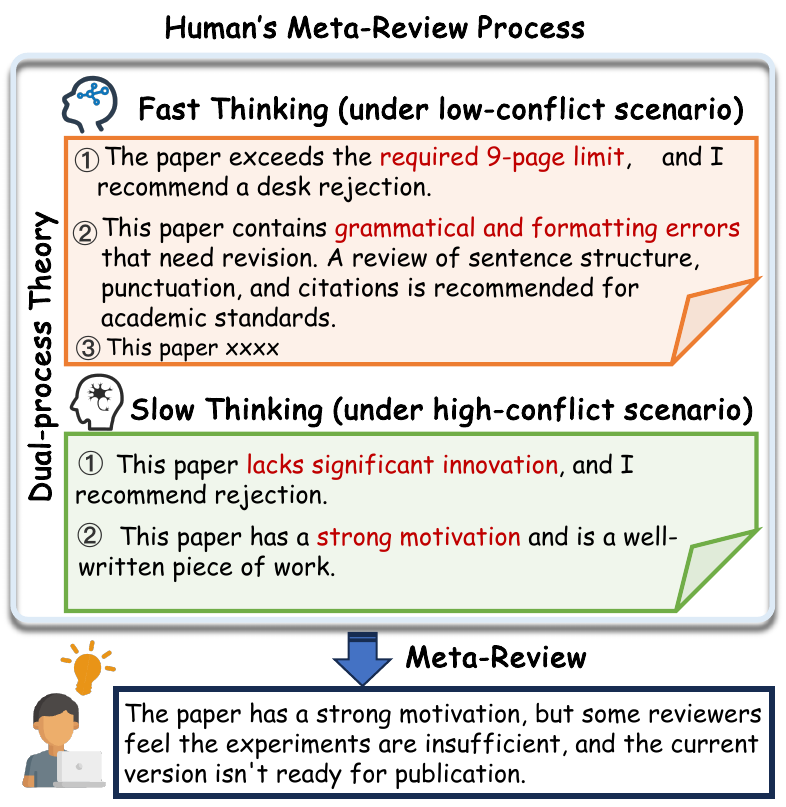}
	}
	\caption{
		Illustrating the ``Fast Thinking” and ``Slow Thinking” processes  in human meta-review.}
	\label{intro}
\end{figure}	

While current LLM-driven review automation mainly targets the peer review process, particularly the generation of detailed comments for individual manuscripts, a crucial yet underexplored aspect is meta-review generation \cite{bhatia2020metagen,santu2024prompting}. This higher-order task involves integrating conflicting evaluations, and deriving consensus through evidence-based reasoning~\cite{li2023summarizing,jin2024agentreview}. Unlike standard summarization, effective meta-review generation requires conflict-aware reasoning that mirrors human decision-making, balancing psychological heuristics and logical deduction \cite{katsikopoulos2011psychological}.

Recently, a few studies \cite{li2024sentiment, zeng2024scientific} have explored simulating human deliberation in meta-review generation, primarily leveraging the Chain-of-Thought (CoT) reasoning framework \cite{wei2022chain}. These approaches decompose the meta-review process into sequential steps: extracting evidence-based opinions, summarizing areas of consensus and controversy, and composing an AC/REJ meta-review. 
However, the absence of cognitive conflict modeling makes these systems prone to biased decision-making, especially when dealing with opposing opinions in peer reviews.
Additionally, we observe that current LLM-based methods unintentionally inherit, and at times amplify, fundamental cognitive biases from human decision-making. These patterns are specifically reflected in:
 \ding{182} Anchoring Effect \cite{wilson1996new}: 
Initial reviewer comments exhibit a stronger anchoring effect on final decisions, with LLMs exhibiting an anchoring coefficient of 0.255, higher than the 0.193 observed in humans.
\ding{183} Conformity Bias \cite{allen1968social}: Majority opinions tend to suppress minority perspectives through pseudo-consensus reinforcement.
The conformity coefficient for LLMs is 0.25, significantly lower than that for humans (1.00), indicating that LLMs are more susceptible to conformity bias.

To bridge this gap, we draw inspiration from dual-process theory~\cite{groves1970habituation,chaiken1999dual}, which posits two distinct modes of cognitive processing: “Fast Thinking” and “Slow Thinking”. As illustrated in Figure~\ref{intro}, the “Fast Think” quickly handles low-conflict cases with immediate responses, while the “Slow Thinking” is engaged for  high-conflict decisions, requiring deeper analysis and reflection. 
Building on this idea, we propose the Cognitive Alignment Framework (CAF) for meta-review generation, which reimagines the role of LLMs in managing multi-perspective discourse in peer review. Specifically, the CAF architecture involves three key phases:
1) Initialization Phase, where the model extracts key points from each comment;
2) Incremental Integration Phase, which progressively incorporates subsequent reviews while maintaining a conflict map; and
3) Cognitive Alignment Phase, which employs dual-process reasoning to balance heuristic pattern recognition with analytical evaluation.
By introducing this adaptive arbitration mechanism, CAF extends the capabilities of LLMs beyond traditional summarization, enabling them to systematically reconcile conflicting scientific opinions. 
%Experiments demonstrate the effectiveness of CAF across multiple LLM-based models. 

In summary, our contributions are threefold:
\begin{itemize}[leftmargin=*,labelindent=0pt,topsep=2pt,itemsep=1pt] 
	\item We uncover inherent cognitive biases, such as the anchoring effect and conformity bias, within existing LLM-driven meta-review systems. To our knowledge, this is the first instance in which these biases have been identified in this task.
	\item We propose a novel three-phase framework that integrates dual-process theory into LLMs. This framework combines iterative conflict mapping and dual reasoning pathways to mitigate biases and resolve scientific disputes.
	\item Experiments demonstrate that CAF outperforms conventional methods, with sentiment consistency gains reaching up to 19.47\% and content consistency improving by as much as 12.95\%.
\end{itemize}

\section{Task Formulation}
%Given a manuscript $\mathcal{P}$ and a set of $N$ peer reviews $\mathcal{R} = \{R_1, \ldots, R_n\}$, where $R_i$ consists of numerical scores $s_i$ and textual comments $c_i$. The objective is to develop an automated system that produces both an editorial decision and a consensus-driven meta-review. It can be formalized as:\begin{equation}	(\mathcal{E}^*, \mathcal{M}^*) = \mathcal{F}\left(\{c_i,s_i\}_{i=1}^N\right)	\tag{1}\end{equation}where $\mathcal{E}^* \in \{\text{Accept}, \text{Reject}\}$ is the final editorial decision, either accepting or rejecting the manuscript. $\mathcal{M}^*$ is the generated meta-review text synthesizing the key arguments from the reviews.$\mathcal{F}$ represents LLM-based framework that processes the peer reviews to generate the editorial decision and meta-review.
Given a set of papers \(\mathcal{A} = \{A_1, A_2, \ldots, A_V\}\), where each paper \(A_i\) is accompanied by a set of \(N\) peer reviews \(\mathcal{R} = \{R_1, \ldots, R_N\}\), and each review \(R_i\) consists of a numerical score \(s_i\) and textual comments \(c_i\), the goal is to develop an automated system that generates both an editorial decision and a consensus-driven meta-review for each paper. This problem can be formalized as follows:
%Given a manuscript $\mathcal{A}$ and a set of $N$ peer reviews $\mathcal{R} = \{(R_1, \ldots, R_N)\}$, where each review $R_i$ consists of a numerical score $s_i$ and textual comments $c_i$, the goal is to develop an automated system that generates both an editorial decision and a consensus-driven meta-review. This can be formalized as:
\begin{equation}
(\mathcal{E}, \mathcal{M}) = \mathcal{F}\left(\{(s_i, c_i)\}_{i=1}^N\right),
\tag{1}
\end{equation}
where
 $\mathcal{E} \in \{\text{Accept}, \text{Reject}\}$ is the final editorial decision\footnote{In this paper, we focus on meta-reviews for conferences in the computer science domain.}, which could either accept or reject the manuscript,
 $\mathcal{M}$ is the generated meta-review text that synthesizes the key arguments from the reviews,
 $\mathcal{F}$ is an LLM-based framework that processes the peer reviews to generate the editorial decision $\mathcal{E}$ and meta-review $\mathcal{M}$.

	\section{Cognitive Biases in LLM-based Meta-Review Generation}
	\label{bias}
	In this section, we will further analyze the existing challenges in current LLM-based meta-review generation systems, particularly focusing on how they may unintentionally introduce human cognitive biases: anchoring effect and conformity bias. 
	 %Although LLMs have made significant strides in automating meta-review generation, it is crucial to recognize that these systems can inadvertently distort the fairness of meta-reviews. In the following experiments, we will be the first to reveal the impact of the anchoring effect and group biases in the meta-review generation task.

%Although LLMs have made substantial progress in automating meta-review generation,  it is important to recognize that current LLM-based systems may unintentionally introduce human cognitive biases, which can distort the fairness of the meta-reviews. In following experiments, we will be the first to reveal the impact of the anchoring effect and group biases in the meta-review generation task.

%To our knowledge, this is the first time that anchoring effect and group biases are identified in the meta-review generation task, and we demonstrate their effects through experiments.

%In this paper, we point out, for the first time, these cognitive biases in meta-review generation and illustrate their effects through experiments.
	
\subsection{Anchoring Effect in Sequential Processing}
		\begin{figure}[t]
	\centering
	\setlength{\fboxrule}{0.pt}
	\setlength{\fboxsep}{0.pt}
	\fbox{
		\includegraphics[width=0.95\linewidth]{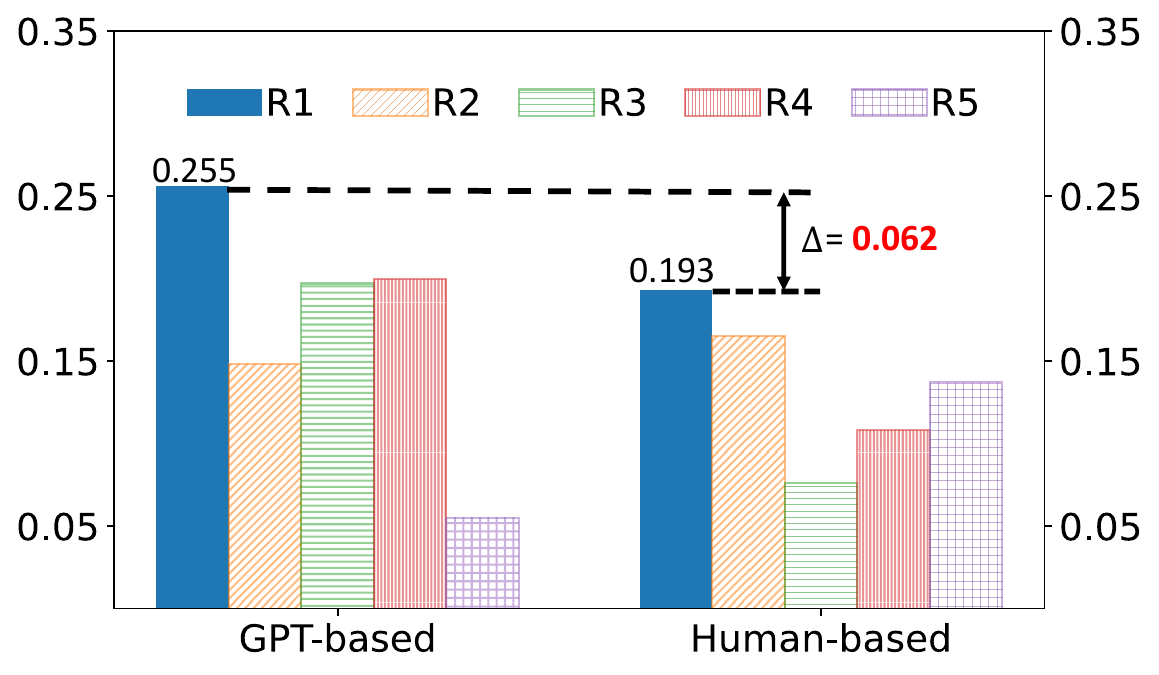}
	}
	\caption{
		The impact of different reviewers (R1, R2, R3, R4, R5) on the final score, with higher y-values indicating greater weight. $\Delta$ represents the difference in anchoring effect between GPT and human reviewers.}
	\label{position_bias}
	\vspace{-10pt}
\end{figure}	
The anchoring effect~\cite{mussweiler2002malleability} refers to the cognitive bias where individuals' judgments are disproportionately influenced by an initial reference point, or ``anchor” during decision-making processes. In meta-review generation, LLMs may exhibit this bias when they first process an initial set of comments or ratings, which can then disproportionately affect their final assessment.

\textbf{Measure.}  
To assess the impact of different reviewers' opinions on final decision, we model this process with logistic regression. Formally, the final score \( \mathcal{E} \) is modeled as a function of a weighted sum of the individual reviewer scores \( s_i \), passed through a Sigmoid function to produce a probability:
\begin{equation}
P(\mathcal{E} = 1 \mid s_1, \dots, s_N) = \sigma\left(\sum_{i=1}^{N} \xi_i \cdot s_i \right),
\end{equation}
%To assess the impact of different reviewers' opinions on the final decision, we model this process as a Logistic regression.This  allows us to quantify the influence of each reviewer's score on overall outcome. Formally, the score \( \mathcal{E} \) is defined as a weighted sum of the individual scores $s_i$
%\begin{equation	\mathcal{E} = \sigma\left( \sum_{i=1}^{N} \xi_i \cdot s_i \right),\end{equation}
where \( \xi_i \) represents the anchoring coefficient for the \( i \)-th review, $\sigma$ represents the sigmoid function, and \( s_i \) is the score provided by reviewer \( i \). Ideally, the coefficients \( \xi_i \) should be uniform to ensure equal representation of all reviews. However, due to the anchoring effect, the coefficient \( \xi_1 \) (for the first review) is often disproportionately emphasized, particularly in CoT methods. By comparing the \( \xi_1 \) values across experiments, we can gauge the strength of the anchoring bias in LLM-based meta-review generation. A higher \( \xi_1 \) value indicates a stronger bias, meaning the first review has had a greater impact on the final outcome.

\textbf{Observation Results.}
We selected 50 peer-reviewed papers from PeerSum benchmark and applied least squares regression to calculate the anchoring coefficients for the different peer review opinions, as shown in Figure~\ref{position_bias}. The results indicate that the recommendations of the first reviewer have a significant influence on the final decision. Specifically, the anchoring coefficient for the first reviewer is 0.255 for GPT-4o, which is substantially higher than the human committee baseline (\( \xi_1 = 0.193 \)). This bias persists even when subsequent reviews provide contradictory evidence, as models tend to overemphasize earlier tokens in their self-attention layers.  This demonstrates that LLMs are particularly sensitive to the first review, and their reliance on earlier inputs can significantly distort the final decision, even in the face of later evidence.

%Experimental results presented in Figure~\ref{position_bias}, across 50 simulated peer review scenarios, show that the recommendations of the first reviewer significantly influence the final decisions. Specifically, anchoring coefficient for the first reviewer are 0.255 for GPT-4o, which is substantially higher than the human committee baseline (\( \xi_1 = 0.193 \)). This bias persists even when subsequent reviews provide contradictory evidence, as models tend to overemphasize earlier tokens in their seolf-attention layers. For example, when initial reviews expressed negative sentiment, only a few of LLM-generated meta-reviews reversed provisional rejections, despite strong contradictory evidence in later reviews \cite{ye2024we}.

%Experimental results in Figure~\ref{position_bias} across 50 simulated peer review scenarios demonstrate that first reviewers' recommendations anchor final decisions with positional bias coefficients ($\Phi_1$) of 0.255 (\textsc{GPT3.5}), significantly exceeding human committee baselines ($\Phi_1=0.193$, $p<0.001$). This effect persists even when subsequent reviews present contradictory evidence, as models disproportionately weight early tokens in their self-attention layers. For instance, when initial reviews contained negative sentiment, only 23\% of LLM-generated meta-reviews reversed provisional rejections despite strong contradictory evidence in later reviews.  
	\subsection{Conformity Bias  via Pseudo-Consensus}  
	\label{Conformity Bia}
Conformity bias, or groupthink \cite{moscovici1972social}, occurs when individuals align their judgments with the majority, even when those judgments differ from their own independent assessments. In LLM-based meta-review generation, this bias arises when the model's output is disproportionately influenced by the majority of prior reviews, rather than synthesizing feedback in an unbiased manner. As a result, the meta-review reflects the dominant opinions, even when they may not be the most accurate or fair assessment.
		\begin{table}[t]
		\renewcommand{\arraystretch}{1.3}
		\resizebox{0.48\textwidth}{!}{
			\begin{tabular}{cccc}
				\hline
				\textbf{Model} & \textbf{Human-based } & \textbf{GPT-3.5}  & \textbf{Qwen2.5-7B} \\
				\hline 		\hline
				$\kappa$ & 1.00  & 0.1250 & 0.2500  \\ 
				Change & -  &     $\downarrow$ 0.8750 & $\downarrow$ 0.7500  \\
				\hline
		\end{tabular}}
		\caption{Conformity coefficient in different LLMs.}
		\label{Conformity coefficient}
	\end{table}
	
	\textbf{Measure.} To quantify the conformity bias, we introduce the conformity coefficient \( \kappa \). Formally, we define a paper set \( \mathcal{A} = \{A_1, A_2, \dots, A_V\} \). For each paper \( A_i \), given \( N \) peer reviews \( \mathcal{R} = \{R_1, R_2, \dots, R_N\} \), we define a function \( V(A_i) \in \{0,1\} \) that aggregates the reviews by majority vote: \( V(A_i) = \text{majority}(R_1, R_2, \dots, R_N) \), where 0 represents rejection and 1 represents acceptance.
	
	Additionally, \( T(A_i) \) represents the ground truth result for paper \( A_i \) (either 0 or 1), while \( F(A_i) \) denotes the predicted result generated by LLMs. We then define the sets \( \Phi \) and \( \Psi \) as follows: 
\begin{equation}
	\begin{aligned}
		\Phi &= \{ A_i \mid V(A_i) = T(A_i) \}, \\
		\Psi &= \{ A_i \mid V(A_i) = F(A_i) \},
	\end{aligned}
\end{equation}
	where \( \Phi \) represents the set of meta-reviews where the majority vote matches the actual outcome, and \( \Psi \) represents the set of predicted meta-reviews that align with the majority vote.
	The value of \( \kappa \) is then given by the proportion of reviews in \( \{\Phi \cap \Psi \} \) relative to the total number of reviews in \( \Psi \):
\begin{equation}
	\kappa = \frac{|\Phi \cap \Psi|}{|\Psi|},
\end{equation}
where\( |\Phi \cap \Psi| \) is the number of meta-reviews where the majority vote matches both the actual outcome and the predicting outcome. This conformity coefficient \( \kappa \) quantifies how much of the decision dominated by the majority opinion in the meta-review is correct.
	Note that a higher value of \( \kappa \) indicates less conformity bias, suggesting the predicted majority vote aligns more closely with  ground truth.

	\textbf{Observation Results.} 
Similarly, we conducted preliminary experiments on 50 peer-reviewed papers from the PeerSum benchmark in Table~\ref{Conformity coefficient}, evaluating two LLM models: Qwen2.5-7B and GPT-3.5, using the indicator $\kappa $ to assess conformity bias. The results revealed that Qwen2.5-7B exhibited a moderate conformity bias with \( \kappa = 0.25 \), while GPT-3.5 showed an even stronger conformity bias at \( \kappa = 0.125 \). Both models demonstrated significant reductions in conformity bias compared to the baseline value of 1.00. These findings highlight the susceptibility of LLMs to conformity bias, with variations across different models. This indicator proved effective in quantifying this bias, and future work will focus on strategies to mitigate it.

		\begin{figure*}[t]
			\centering
			\setlength{\fboxrule}{0.pt}
			\setlength{\fboxsep}{0.pt}
			\fbox{
				\includegraphics[width=0.99\linewidth]{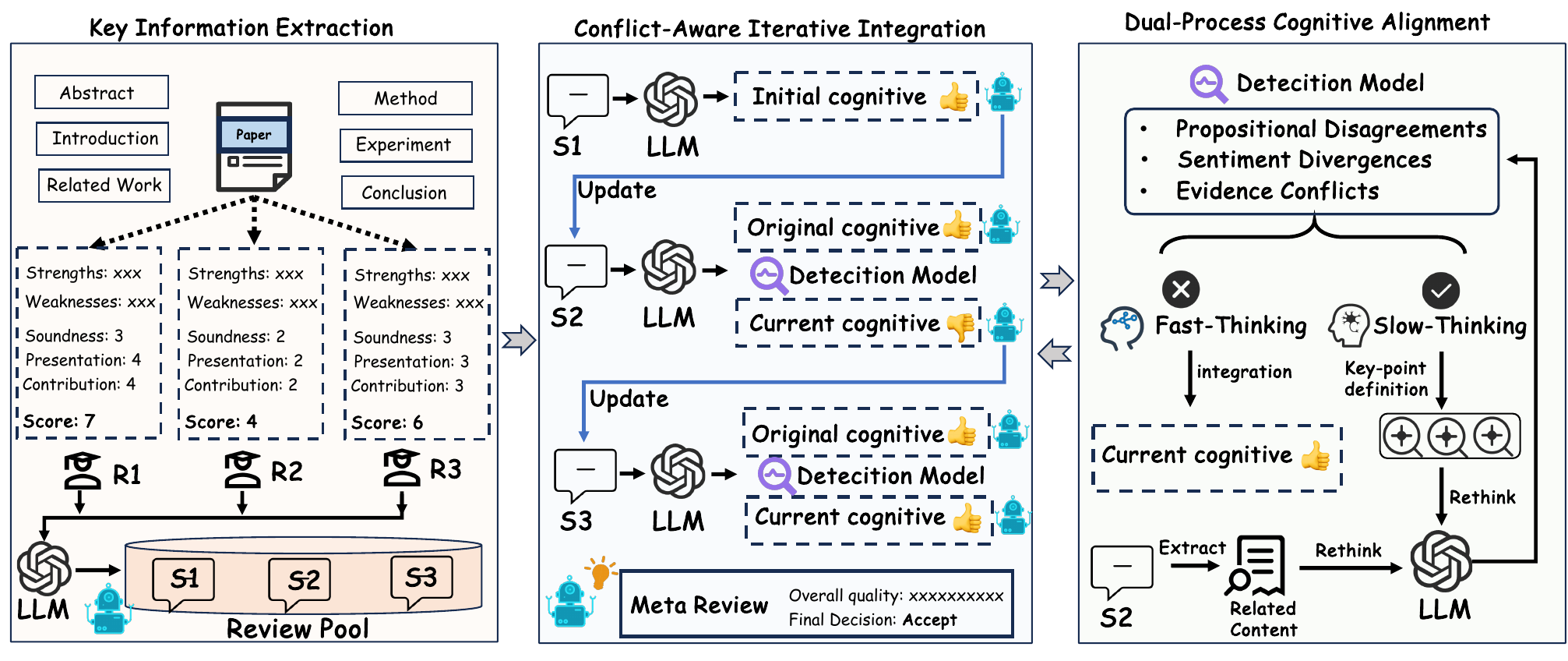}
			}
			\caption{
			The architecture of our proposed CAF model. It begins with Key Information Extraction, where initial reviews are collected and key elements are extracted ($\S \ref{s1}$). Next, the Conflict-Aware Iterative Integration phase integrates the reviews while detecting and resolving conflicts ($\S \ref{s2}$). Finally, the Dual-Process Cognitive Alignment stage applies both fast and slow thinking strategies to address conflicts, ultimately generating the meta-review ($\S \ref{s3}$).}
			\label{model}
		\end{figure*}			
\section{Methodology}
Existing methods fail to effectively handle conflicting viewpoints within differing opinions, and often introduce additional cognitive biases, such as anchoring effects and conformity bias.
To overcome these limitations, we propose a novel framework CAF in Figure~\ref{model}. It implements a three-phase cognitive architecture for meta-review synthesis, inspired by dual-process cognitive theory. 
%As shown in Figure~\ref{model}, the system alternates between rapid integration and deliberative conflict resolution.
%To address these challenges, we propose a novel framework, CAF, which implements a three-phase cognitive architecture for meta-review synthesis, drawing on dual-process cognitive theory. As illustrated in Figure~\ref{model}, the system alternates between rapid integration and thoughtful conflict resolution.

\subsection{Key Information Extraction}
\label{s1}
For a target paper with $N$ reviews $\mathcal{R} = \{R_1, R_2, \dots, R_N\}$, we establish the foundational context $\mathcal{S}$ through structured knowledge extraction from all initial reviews. Since review texts can be lengthy and redundant, processing them in full may lead to inefficiencies. To mitigate this, we first extract key elements from each individual review:
\begin{equation}
	\mathcal{S}_{1 \sim N} = \bigcup_{i=1}^{N} \mathcal{F}_{\text{LLM}}(R_i),
	\label{s}
\end{equation}
where $\mathcal{F}_{\text{LLM}}$ is a LLM that identifies and summarizes essential aspects from each review. By extracting these critical points, the synthesis process is streamlined, avoiding information overload while retaining relevant insights. The extracted elements are then organized into a preliminary structure to facilitate iterative refinement. The initial prompt used is provided in Appendix \ref{Init_prompt}.

\subsection{Conflict-Aware Iterative Integration}
\label{s2}
%After initializing the review pool, the framework incrementally synthesizes reviews while simultaneously identifying inconsistencies. Since the meta-review $\mathcal{M}$ is initially empty when the first reviewer’s input $\mathcal{S}_1$ is received, it is directly assigned as the preliminary meta-review: $\mathcal{M}_1 = \mathcal{S}_1$.This assignment establishes a foundational context for integrating subsequent reviews.
After initializing the review pool, the framework incrementally synthesizes reviews while simultaneously identifying inconsistencies. Traditional LLM-based methods~\cite{zeng2024scientific,li2024sentiment}, which process all reviews at once, can exacerbate issues like anchoring effects and conformity bias. Specifically, when all reviews are considered together, early reviews often serve as anchors, disproportionately influencing the synthesis process and leading to biased conclusions. Similarly, the presence of multiple reviews at once can cause the model to align too closely with the majority opinion, reinforcing conformity bias. Our approach mitigates these biases by incorporating reviews progressively, allowing the system to adjust its synthesis in response to each new input.

%After initializing the review pool, the framework incrementally synthesizes reviews while simultaneously identifying inconsistencies. When the first reviewer's input $\mathcal{S}_1$ is received, the initial cognitive $\mathcal{C}_{1}$ of the meta-review is fully derived from the first review. This assignment establishes a foundational cognitive for integrating subsequent reviews.
Initially, when the first reviewer's input \( \mathcal{S}_1 \) is received, the initial cognitive representation \( \mathcal{C}_1 \) of the meta-review is fully derived from the first review. 
Subsequently, when the second reviewer's input \( \mathcal{S}_2 \) arrives, the framework updates the existing cognitive iteratively. Given the current cognitive context \( \mathcal{C}_{i-1} \) and the incoming review \( \mathcal{S}_i \), a conflict detection model \( \mathcal{D} \) assesses potential inconsistencies:
\begin{equation}
	\mathcal{D}(\mathcal{C}_{i-1}, \mathcal{S}_i) =
	\begin{cases}
		1, & \text{if a conflict exists,} \\
		0, & \text{if no conflict exists.}
	\end{cases}
\end{equation}
This module identifies inconsistencies based on three key factors:
\ding{182} Propositional Disagreements: Contradictions in claims or assessments of the paper’s methodology and contributions,
\ding{183} Sentiment Divergences: Variations in sentiment across reviewers, such as one being highly positive while another is critical,
\ding{184} Evidence Conflicts: Differences in how reviewers interpret and weigh empirical evidence.
This conflict-aware integration process ensures that all perspectives are carefully considered, progressively refining the meta-review while addressing inconsistencies systematically. 

This continues until all reviews (i.e., $\mathcal{S}_3$, \dots, $\mathcal{S}_N$) have been integrated into the model.

%After initializing the review pool, the framework begins the incremental synthesis of reviews while identifying inconsistencies. When the first reviewer’s input $R_1$ is received, the meta-review $M$ is initially empty. As a result, the first review is directly assigned as the preliminary meta-review: $M = R_1$.

\subsection{Dual-Process Cognitive Alignment}
\label{s3}
In each conflict detection process, the CAF performs a cognitive alignment procedure, balancing rapid heuristic processing with analytical reasoning. If no conflict is detected, the system directly integrates \( \mathcal{S}_i \) into the current cognitive framework. However, if a conflict is identified, a structured conflict resolution mechanism is triggered to maintain coherence and consistency within the meta-review. This cycle of conflict detection continues until no conflict remains or the predefined maximum number $Q$ of iterations is reached. During this phase, the model engages in cognitive reconstruction, generating a refined set of essential arguments, known as the \textit{key-point definition} \( \mathcal{K}_i \).
%In this step, our framework follows a dual-process model inspired by human cognition, balancing rapid heuristic processing (fast think) and deliberate analytical reasoning (slow think). If no conflict is detected, the system proceeds with the direct integration of $\mathcal{S}_i$ into the synthesized context. However, if a conflict is identified, a structured conflict resolution mechanism is triggered to ensure coherence and consistency within the meta-review.During this conflict resolution phase, the model engages in a cognitive reconstruction procedure, where it generates a refined set of essential arguments, termed the \textit{key-point definition} $\mathcal{K}_i$. 
This two processes (i.e., fast thinking and slow thinking) aims to integrate conflicting points into a unified position and are formally expressed as follows:
\begin{equation}
	\mathcal{K}_i = \mathcal{G}_{\text{Fast}}(\mathcal{C}_{i-1}, \mathcal{S}_i) \oplus \mathcal{G}_{\text{Slow}}(\mathcal{C}_{i-1}, \mathcal{S}_i),
\end{equation}
where ``$\oplus$”denotes concatenation fusion. Here, $\mathcal{G}_{\text{Fast}}$ facilitates the rapid identification of argument structures and sentiment trends, while $\mathcal{G}_{\text{Slow}}$ serves as a reconciliation function, guiding the model to rethink and generate a cohesive set of key points. This process helps minimize contradictions and enhances the overall meta-review synthesis.
The refined cognitive is then updated as follows: \begin{equation} \mathcal{C}_i = \mathcal{U}(\mathcal{C}_{i-1}, \mathcal{K}_i, \mathcal{S}_i), \end{equation} where $\mathcal{U}$ integrates the resolved key-points into the current cognitive.
After all reviews have been processed and conflicts resolved, the final meta-review is given by \( \mathcal{M}_{\text{final}} = f(\mathcal{C}_N) \), where \( f \) is a synthesis function based on the current cognitive knowledge \( \mathcal{C}_N \). This structured meta-review provides a balanced, evidence-supported summary, ensuring that disagreements are systematically addressed while preserving key insights from all peer reviews. 

The entire process is outlined in Algorithm \ref{algorithm1}.

\begin{algorithm}[t]
	\setstretch{0.8}
	\caption{The three-phase cognitive framework for meta-review synthesis}
		\label{algorithm1}
	\KwIn{$\mathcal{R} = \{R_1, R_2, \dots, R_N\}$: Reviews}
	\KwOut{$\mathcal{M}$: Synthesized Meta-Review}
	
	\textbf{Step I: Review Pool Initialization} \\
	Initialization review $\mathcal{S}$ via Eq. (\ref{s}) \\
	Initialization meta-review $\mathcal{M} = \varnothing$
	
	\For{$i = 1 \textbf{ to } N$}{
		\textbf{Step II: Iterative Integration} \\
		\If{$i == 1$}{
			$\mathcal{C}_1 \gets \mathcal{F}_{\text{LLM}}(R_1)$ \\
		}
		\Else{
			\textbf{Step III: Cognitive Alignment} \\
			$conflict \gets \mathcal{D}(\mathcal{C}_{i-1}, \mathcal{S}_i)$ \\
			Initialize $T \gets 0$ \\
			\While{$conflict == 1 \land T < Q$}{
				$\mathcal{K}_i \gets \mathcal{G}_{\text{Slow}}(\mathcal{C}_{i-1}, \mathcal{S}_i)$ \\
				$\mathcal{C}_i \gets \mathcal{U}(\mathcal{C}_{i-1}, \mathcal{K}_i, \mathcal{S}_i)$ \\
				$conflict \gets \mathcal{D}(\mathcal{C}_{i-1}, \mathcal{S}_i)$ \\
				$ T\gets T + 1$ \\
			}
				$\mathcal{K}_i \gets \mathcal{G}_{\text{Fast}}(\mathcal{C}_{i-1}, \mathcal{S}_i)$ \\
				$\mathcal{C}_i \gets \mathcal{U}(\mathcal{C}_{i-1}, \mathcal{K}_i)$ \\

		}
	}	Generate the final meta-review $\mathcal{M} = f(\mathcal{C}_N)$ \\
	\Return{$\mathcal{M}$}
\end{algorithm}

\begin{table*}[t]
	\renewcommand{\arraystretch}{1.05}
	\resizebox{1\textwidth}{!}{
		\centering
		\begin{tabular}{ccccccccr}
			\hline
			LLMs & Evaluation Metric & Prompt-Naive & Prompt-Increm. & Prompt-SelfCoT & Prompt-SCF & Prompt-CAF & Improve.  \\
			\hline 		\hline
			\multirow{4}{*}{GPT-3.5} 
			& FacetEval & 0.3751 & 0.3909 & 0.3653 & \underline{0.4788} & \textbf{0.5551} & {\textcolor{purple}{$\blacktriangle$\!\!} 15.93\%}\\
			& ROUGE-1 & 0.2873 & \underline{0.2930} & 0.2760 & 0.2806 & \textbf{0.2987}& {\textcolor{purple}{$\blacktriangle$~}   1.94\%} \\
			& ROUGE-2 & 0.0779 & \underline{0.0859} & 0.0662 & 0.0738 & \textbf{0.0898} &{\textcolor{purple}{$\blacktriangle$~} 4.54\%}\\
			& ROUGE-L & 0.2177 & \underline{0.2292} & 0.2089 & 0.2134 & \textbf{0.2343}& {\textcolor{purple}{$\blacktriangle$~} 2.22\%} \\
			\hline
			\multirow{4}{*}{GPT-4o} 
			& FacetEval  &0.4359 & 0.4037 & 0.3956 & \textbf{0.5898} & \underline{0.5582}& {\textcolor{purple}{$\blacktriangledown$~} 5.35\%} \\
			& ROUGE-1 & \underline{0.2965} & 0.2907 & 0.2625 & 0.2896 & \textbf{0.3099} & {\textcolor{purple}{$\blacktriangle$~} 4.51\%}\\
			& ROUGE-2 & \underline{0.0934} & 0.0893 & 0.0629 & 0.0868 & \textbf{0.1055}& {\textcolor{purple}{$\blacktriangle$\!\!} 12.95\%} \\
			& ROUGE-L & \underline{0.2338} & \underline{0.2338} & 0.2091 & 0.2248 & \textbf{0.2553} & {\textcolor{purple}{$\blacktriangle$~} 9.19\%}\\
			\hline
			\multirow{4}{*}{Qwen2.5-7B} 
			& FacetEval & 0.5082 & 0.4929 & 0.4925 & \textbf{0.5693} & \underline{0.5690} & {\textcolor{purple}{$\blacktriangledown$~} 0.05\%}\\
			& ROUGE-1 & 0.2669 & \underline{0.2677} & 0.2674 & 0.2629 & \textbf{0.2774}& {\textcolor{purple}{$\blacktriangle$~} 3.62\%} \\
			& ROUGE-2 & \underline{0.0748} & 0.0722 & 0.0643 & 0.0710 & \textbf{0.0791} & {\textcolor{purple}{$\blacktriangle$~} 5.74\%}\\
			& ROUGE-L & 0.2069 & 0.2093 & \underline{0.2110} & 0.2036 & \textbf{0.2243} & {\textcolor{purple}{$\blacktriangle$~} 6.30\%}\\
			\hline
			\multirow{4}{*}{Llama3-8B} 
			& FacetEval  & 0.3368 & 0.3522 & 0.3170 & \underline{0.4067} & \textbf{0.4859} & {\textcolor{purple}{$\blacktriangle$\!\!} 19.47\%}\\
			& ROUGE-1 & 0.2753 & \underline{0.2864} & 0.2490 & 0.2800 & \textbf{0.2879} & {\textcolor{purple}{$\blacktriangle$~} 0.52\%}\\
			& ROUGE-2 & 0.0815 & \underline{0.0937} & 0.0519 & 0.0807 & \textbf{0.0957}& {\textcolor{purple}{$\blacktriangle$~} 2.13\%} \\
			& ROUGE-L & 0.2287 & \underline{0.2397} & 0.2075 & 0.2295 & \textbf{0.2420} & {\textcolor{purple}{$\blacktriangle$~} 0.96\%}\\
			\hline
	\end{tabular}}
			\vspace{-5pt}
	\caption{Performances of different LLMs with different prompting methods. The bold and underlined values represent the best and second best in each row, respectively. The red upward triangle indicates an improvement.}
	\label{main}
\end{table*}

				\section{Experiment}
				In this section, we conduct experiments to answer the following research questions:
\textbf{RQ1:} How does the proposed CAF model compare to other LLM-based methods? \textbf{RQ2:} In which scenarios does the fast-slow thinking prompt strategy most impact meta-review generation? \textbf{RQ3:} Does the CAF model mitigate the two cognitive biases found in existing methods?
\textbf{RQ4:} How can we more intuitively showcase the superiority of our method in the meta-review generation task?
				\subsection{Experiment Settings}
				\textbf{Dataset.} Following~\cite{li2023summarizing,li2024sentiment}, we employ PeerSum dataset\footnote{\url{https://github.com/oaimli/PeerSum}} for our experiments. This dataset, created by crawling peer reviews from OpenReview, includes 14,993 submissions from two major machine learning conferences: NeurIPS (2021-2022) and ICLR (2018-2022).
				
				\noindent
				\textbf{Baselines.}
				In this section, we used the following LLM-based approaches:  
				Prompt-Naive,
				Prompt-Increm.,
				Prompt-SelfCoT,
				Prompt-SCF. 
				
				\noindent
				\textbf{Implementation Details.}						
				We conducted experiments with four LLMs: Qwen2.5-7B~\cite{hui2024qwen2}, Llama3-8B \cite{dubey2024llama}, GPT-3.5 \cite{ye2023comprehensive} and GPT-4o \cite{hurst2024gpt}. 
				
				\noindent
				\textbf{Evaluation.} 	
				We evaluate the generated meta-review from two perspectives: sentiment consistency and content consistency. For sentiment consistency, we utilize \texttt{FacetEval}~\cite{li2024sentiment}, while for content consistency, we employ the ROUGE metric \cite{lin2004rouge}, which includes \texttt{ROUGE-1}, \texttt{ROUGE-2}, and \texttt{ROUGE-L}.

More details can be found in Appendix~\ref{appendix_experiment}.

%In terms of evaluation metrics, we evaluate the generated meta-review from two aspects: emotional consistency and content consistency. For sentiment, we use FacetEval from SCF to assess the quality and emotional consistency of the LLM-generated meta-review. We label the sentiment of the ground-truth meta-review using GPT-4o and classify it into five sentiment categories: strongly positive, positive, neutral, negative, and strongly negative. To quantify emotional consistency, we use one-hot encoding to map the sentiment of each aspect to the vector space and concatenate these aspect-level vectors. Then, we calculate the cosine similarity between the generated emotional vector and the ground-truth emotional vector to obtain the emotional consistency score. In terms of content, we use ROUGE to measure text consistency (including ROUGE-1, ROUGE-2, ROUGE-L). ROUGE is a commonly used automatic evaluation metric primarily used to assess the quality of text in natural language generation tasks. It calculates similarity by comparing the overlap between the generated content and the true content. 

														\begin{figure}[t]
							\centering
							\setlength{\fboxrule}{0.pt}
							\setlength{\fboxsep}{0.pt}
							\fbox{
								\includegraphics[width=0.95\linewidth]{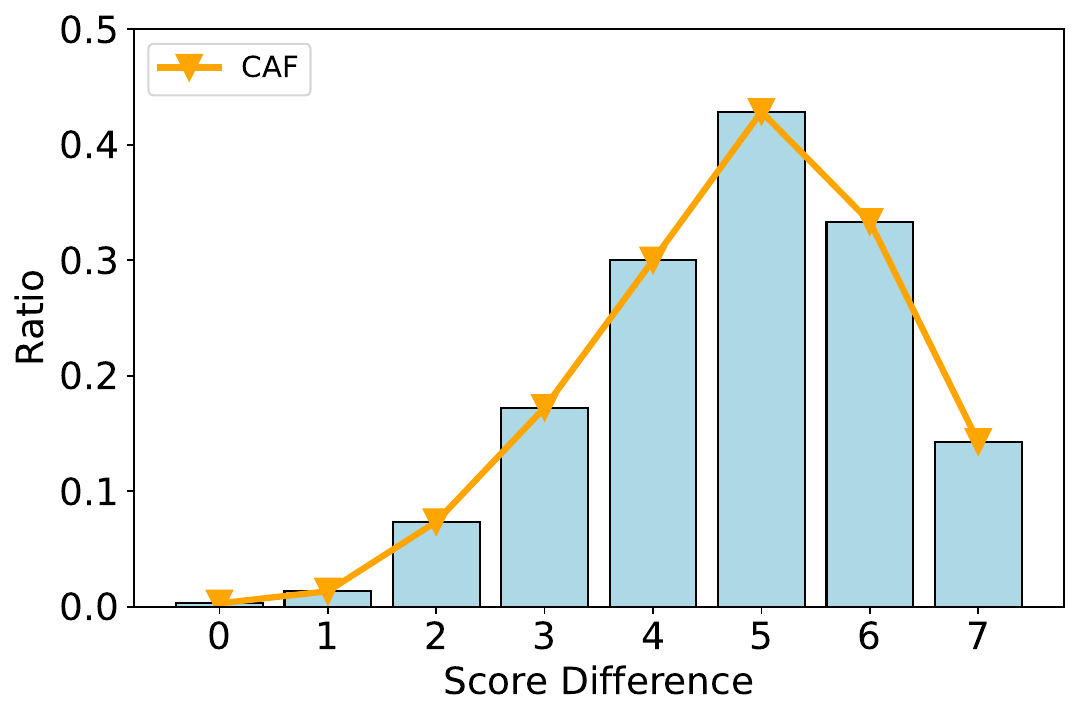}
							}
							\caption{
								The relationship between score differences in reviews and the probability of needing cognitive reconstruction (reflection) in the meta-review process.}
							\label{rethink}
						\end{figure}	
						\subsection{Main Experimental  Results (RQ1)}
						To validate the effectiveness of CAF, we conducted experiments across multiple models in Table~\ref{main}. 
						
						The  results shows that our proposed framework excels in both sentiment consistency and text quality when generating meta-reviews. Specifically, on GPT-3.5 and Llama3-8B, our generated meta-reviews significantly outperform other baseline methods in terms of sentiment consistency and ROUGE scores. By integrating the “Fast-Slow Thinking” theory, CAF effectively balances intuitive  and deliberative strategies,
						resulting in meta-reviews that are fluent, professional, and closely aligned with real-world meta-review standards. 
						
						However, on GPT-4o and Qwen2.5-7B, our method does not surpass the Prompt-SCF approach in sentiment consistency. We hypothesize that this is due to GPT-4o's inherent capabilities, which allow it to achieve high sentiment consistency with relatively simple prompts. Additionally, since SCF explicitly optimizes for sentiment consistency, it performs better in this aspect on GPT-4o.
						Overall, despite the observed limitations in certain cases, our method demonstrates strong generation capabilities across most models, particularly in enhancing sentiment and content consistency. These findings validate the effectiveness of CAF in automatically generating high-quality meta-reviews.

\subsection{Further Analysis (RQ2)}			
To explore under what circumstances our model engages in cognitive reconstruction, we analyzed the relationship between the score differences in comments among reviewers and the probability of reflection for each paper, as shown in Figure \ref{rethink}. 

The results show that as the score difference in opinions between reviewers increases, the ratio of reflection rises, peaking at a difference of 5. This suggests that when there is moderate to large divergence in reviews, more reflection is needed to reconcile conflicting viewpoints and ensure consistency in the final meta-review.
The gradual increase in reflection up to the peak reflects the cognitive process of resolving discrepancies. The subsequent decline indicates that for very small or large differences, less reflection is required. In cases of small differences, meta-reviewers can quickly align their opinions, while for large differences, reflection may decrease as reviewers focus on clear-cut arguments or rely on existing consensus.

		\begin{figure}[t]
	\centering
	\setlength{\fboxrule}{0.pt}
	\setlength{\fboxsep}{0.pt}
	\fbox{
		\includegraphics[width=0.95\linewidth]{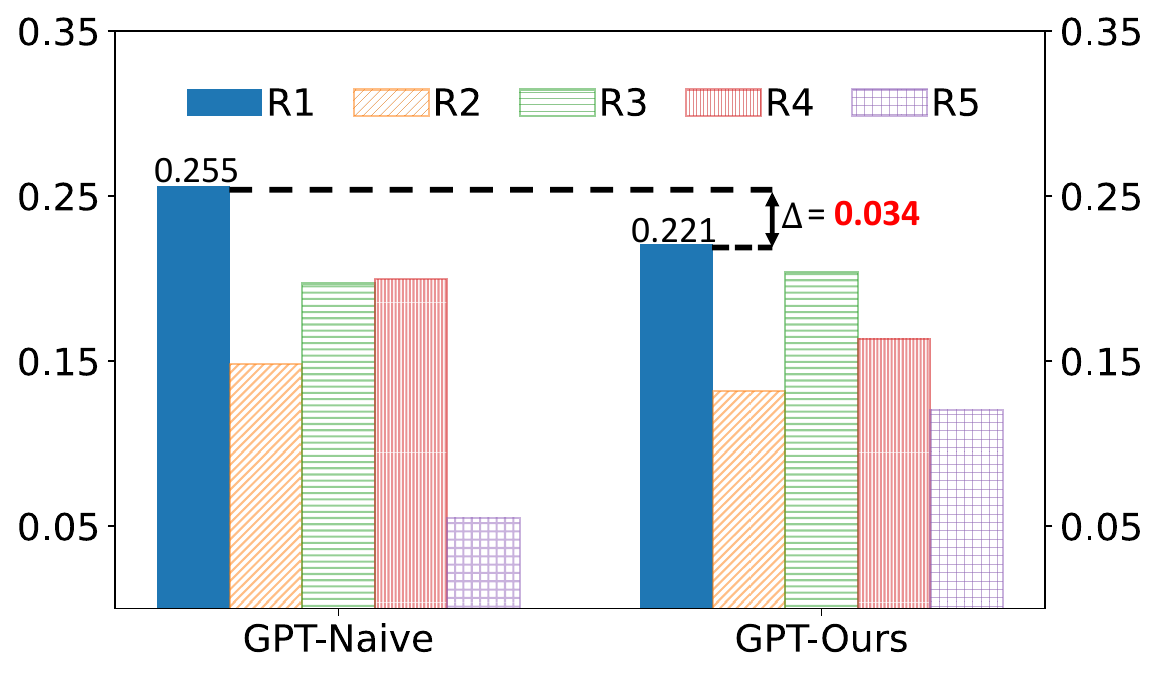}
	}
	\caption{
		The impact of different reviewers on the final score, with higher y-values indicating greater weight.}
	\label{positive_improve}
\end{figure}	
			\begin{table}[t]
			\renewcommand{\arraystretch}{1.3}
			\resizebox{0.48\textwidth}{!}{
				\begin{tabular}{cccc}
					\hline
					\textbf{Model} & \textbf{Prompt-Naive} & \textbf{Prompt-CAF}  & \textbf{Improve} \\
					\hline 		\hline
					GPT-3.5 & 0.1250  & 0.4444 & 0.3194 $\uparrow$  \\
					Qwen2.5-7B & 0.2500  & 0.5000 & 0.2500 $\uparrow$  \\
					\hline
			\end{tabular}}
			\caption{Conformity coefficient in different LLMs.}
			\label{group_improve}
		\end{table}				
						\subsection{Cognitive Bias Mitigation (RQ3)}
In \S \ref{bias}, we discussed how existing methods can lead to the anchoring effect and conformity bias. To evaluate whether CAF mitigates these biases, we conducted further experiments in this section.

\noindent
\textbf{Primacy Effect.}
As shown in \ref{positive_improve}, the GPT-Naive model exhibits a higher primacy effect (0.255) compared to GPT-Ours (0.221), with a difference of 0.034. This indicates that our approach effectively reduces the anchoring effect helping overcome the bias toward earlier information. LLMs are inherently sensitive to earlier inputs due to their self-attention mechanisms, and previous methods process all reviews at once, exacerbating this bias. In contrast, our incremental input approach allows the model to reason with each new review sequentially, reducing the influence of earlier reviews.

\noindent
\textbf{Conformity Bias.}
Table \ref{group_improve} shows that CAF improves evaluation consistency across models. For GPT-3.5, the improvement is 0.3194 (from 0.1250 to 0.4444), and for Qwen2.5-7B, it is 0.25 (from 0.25 to 0.5). These results demonstrate that our method reduces conformity bias, resulting in more independent and diverse outputs.
Existing methods process all reviews at once, making the model more susceptible to conformity bias as it aligns with the majority opinion. In contrast, our incremental input approach, combined with fast-slow thinking for reasoning, allows the model to reprocess each review iteratively, making it less likely to be swayed by dominant viewpoints.

%Overall, our approach effectively mitigates both the primacy and conformity biases, offering a more balanced framework for model evaluation.
						
\subsection{Case Study (RQ4)}
To better illustrate the effectiveness of CAF, we conducted a case study shown in Figure \ref{case}. The complete case study can be found in Appendix \ref{case_study}.

In this case, both the first and second reviewers praised the paper. However, the third reviewer raised some concerns regarding the methodology. Upon detecting this conflict, CAF engaged in a “Slow Thinking” process, analyzing the feedback and identifying four key issues: Planckian Jitter, Test Dataset Manipulation, Image Resolution, and Gamma Correction. After reflecting on these issues, the conflict was resolved, leading to the paper's rejection due to unresolved issues and lack of demonstrated validity. This decision aligned with the ground truth of the meta-review.
Naive prompts, on the other hand, fail to identify such conflicts. While the naive meta-review pointed out some weaknesses, it missed the critical issue of experimental validity and favored the majority opinion, recommending acceptance.

		\begin{figure*}[t]
	\centering
	\setlength{\fboxrule}{0.pt}
	\setlength{\fboxsep}{0.pt}
	\fbox{
		\includegraphics[width=1\linewidth]{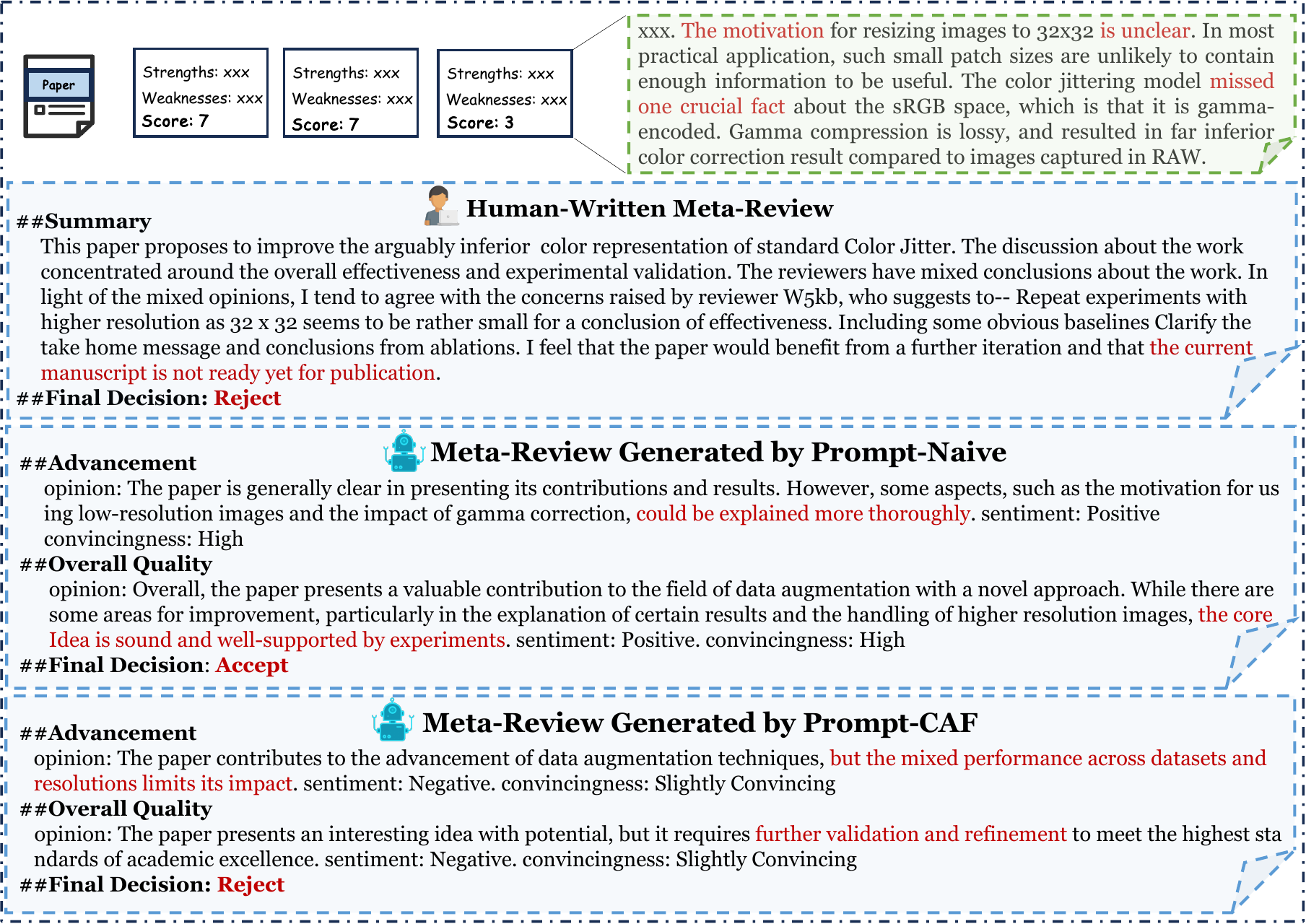}
	}
\caption{Meta-reviews generated by different methods, with red sections highlighting core viewpoints.}
	\label{case}
\end{figure*}	
	\section{Related Work}
\textbf{LLM-powered Peer Review.}
Recently, Large Language Models (LLMs) have made remarkable advancements in text generation tasks~\cite{zhao2023survey, luo2024let}, inspiring researchers to explore their potential in enhancing the peer review process~\cite{yu2024your, jin2024agentreview}. A major area of focus has been the automation of individual reviews to enhance the quality of academic assessments. For example, \cite{liu2023reviewergpt} and \cite{liang2024can} have developed custom prompts for guiding GPT-4 in generating specific scientific feedback. Meanwhile, \cite{wei2023academicgpt} have explored the use of continuous pretraining to better align LLMs with human-like review behaviors, and \cite{jin2024agentreview} have leveraged LLMs to simulate the entire review process, examining how different factors impact academic evaluation.
However, these studies mainly focus on the peer review itself and overlook the crucial aspect of meta-reviewing, which involves synthesizing feedback from multiple reviewers \cite{li2023summarizing,li2024sentiment}. 
\noindent
\textbf{Applications of Dual-process Theory.}
Dual-process theory~\cite{groves1970habituation,chaiken1999dual} provides a psychological framework for understanding human thinking and decision-making, proposing two distinct modes of cognitive processing: fast, automatic thinking and slow, deliberate reasoning. This theory has been effectively applied to machine learning approaches. For example, \cite{mittal2017thinking} conceptualizes the vector space model as representing fast thinking and knowledge graph reasoning as slow thinking. \cite{bengio2017consciousness} highlights the connection between reflective thinking and consciousness, while \cite{chen2019deep} introduces an end-to-end framework integrating a generative decoder for fast thinking with a reasoning module for slow thinking to tackle complex tasks. Building on these ideas, \cite{liu2022neural} combines neural networks and symbolic reasoning to solve question-answering challenges, and \cite{he2024planning} designs a dialogue planning framework using a fast policy language model and a slow Monte Carlo tree search planner.
Inspired by these works, we apply the dual-process theory to simulate the human peer review process.

				\section{Conclusion}
				 In this paper, we propose the Cognitive Alignment Framework (CAF), which leverages dual-process cognitive modeling to enhance meta-review synthesis in LLM-based systems. By incorporating a three-phase processing pipeline, CAF effectively manages conflicting viewpoints, enabling more balanced and coherent meta-reviews.
				 The experimental results validates the effectiveness of our proposed CAF across multiple LLM-based models. 
				 %Experiments show significant improvements in consensus accuracy, demonstrating that CAF not only aggregates information but also resolves scientific disagreements in a human-aligned manner. 
				 %These results highlight the potential of structured cognitive modeling to improve automated meta-review processes, offering a more reliable and efficient approach to academic evaluation.
				\section*{Limitations}
CAF is designed to assist authors in understanding review comments during the peer review process, particularly in the meta-review stage. However, there are several limitations to this approach, as well as opportunities for future exploration.

\noindent
\textbf{Data Scope.} The dataset used in this study is primarily sourced from the PeerSum benchmark, which focuses on machine learning papers. As a result, most of the meta-reviews in the dataset are specific to this domain. This limits the generalizability of the findings, as the conclusions drawn may not apply to other academic fields, and the framework's effectiveness across diverse domains remains untested.

\noindent
\textbf{Rebuttal Exclusion.} For simplicity, the current framework does not incorporate author rebuttals during meta-review generation. Including rebuttal content could enhance the synthesis by considering author responses and alternative perspectives, which may influence the final meta-review. The exclusion of this element may restrict the depth and nuance captured in the generated reviews.
				
\section*{Ethics Statement}
Although large language models (LLMs) have shown promise in generating useful meta-reviews, it is crucial to note that their results should not be relied upon without human oversight. LLMs may occasionally produce inaccurate content or reflect biases present in their training data. We have made efforts to mitigate bias and ensure fairness, though some unintended biases may still occur. Furthermore, we have prioritized data privacy and confidentiality, complying with relevant data protection regulations. Our objective is not to replace human reviewers, but to demonstrate that LLMs can be valuable tools to assist and streamline the meta-review process, enhancing both efficiency and effectiveness when used alongside expert judgment.

	\bibliography{anthology}
				
\newpage
\appendix
	\section{Experiment Settings}
	\label{appendix_experiment}
\subsection{Dataset} Following prior work~\cite{li2023summarizing,li2024sentiment}, we employ Peersum dataset for our experiments. This comprehensive dataset, constructed by crawling peer reviews from OpenReview, aggregates 14,993 submission instances from two major machine learning conferences: NeurIPS (2021-2022) and ICLR (2018-2022). Each data instance contains rich academic discourse elements including reviewer comments with associated numerical scores, public discussion threads, meta-reviews summarizing the discussion, and final decision.We randomly selected 300 papers from this dataset for the main experiment. Given the context length limitation of LLMs, this experiment excludes rebuttal content and public comments, focusing solely on the reviews and rating data from professional reviewers. Additionally, the reviewers are assigned an order based on the list sequence to ensure that the review input order for each method remains consistent throughout the experiment. For the sentiment labeling of real meta reviews, we use GPT-4o to automate the labeling of meta reviews in the dataset for subsequent sentiment consistency evaluation. The specific prompt is provided Appendix \ref{prompt} .

\subsection{Baselines}
In this section, we used the following approaches:  
\ding{182} \textbf{Prompt-Naive}: Generate the meta-review in a single step using a simple query,
\ding{183} \textbf{Prompt-Increm.}: After each incremental review, express opinions and generate the meta-review only after the final review,
\ding{184} \textbf{Prompt-SelfCoT} \cite{zeng2024scientific}: First, prompt the LLM to outline the steps for generating a meta-review, then have it produce the meta-review according to those outlined steps,
\ding{185} \textbf{Prompt-SCF}  \cite{li2024sentiment}: Generate the meta-review using pre-defined, sentiment-aware generation steps.
Please refer to Appendix \ref{prompt} for the detailed baseline prompts.

\subsection{Implementation Details}						
We conducted experiments with two open-source LLMs: Qwen2.5-7B~\cite{hui2024qwen2} and Llama3-8B \cite{dubey2024llama}, and two closed-source LLMs: GPT-3.5 \cite{ye2023comprehensive} and GPT-4o \cite{islam2024gpt}. To ensure the reproducibility of our experiments, we fixed the temperature parameter at 0. Additionally, to avoid the system getting stuck in an infinite reasoning loop, we set the maximum number of conflict resolution attempts to 5. In $\S$ \ref{Conformity Bia}  , we set the acceptance/rejection score threshold at 6, meaning that when a reviewer gives a score of 6 or higher, the reviewer's opinion is considered as accepting the paper, and conversely, as rejecting it. Thus, the scores given by the reviewers can be mapped to acceptance and rejection decisions. The code is provided at	\href{https://anonymous.4open.science/r/CAF\_metareviewing-2A70}{CAF}.

\subsection{Evaluation} 	
We assess the generated meta-review from two main aspects: sentiment consistency and content consistency. 
\ding{182} For sentiment consistency, we use \texttt{FacetEval} from SCF to evaluate the quality and sentiment alignment of the LLM-generated meta-review. To label the sentiment of the ground-truth meta-review, we employ GPT-4 and classify it into five sentiment categories: strongly positive, positive, neutral, negative, and strongly negative. To quantify emotional consistency, we utilize one-hot encoding to map the sentiment of each aspect to a vector space, and then concatenate these aspect-level vectors. Subsequently, we compute the cosine similarity between the generated emotional vector and the ground-truth emotional vector to derive the emotional consistency score.
\ding{183} For content consistency, we use the ROUGE metric, which includes \texttt{ROUGE-1}, \texttt{ROUGE-2}, and \texttt{ROUGE-L}, to evaluate text consistency. ROUGE is a widely adopted automatic evaluation metric primarily used to assess the quality of text in natural language generation tasks. It calculates the similarity by comparing the overlap between the generated content and the ground-truth content.

%In terms of evaluation metrics, we evaluate the generated meta-review from two aspects: emotional consistency and content consistency. For sentiment, we use FacetEval from SCF to assess the quality and emotional consistency of the LLM-generated meta-review. We label the sentiment of the ground-truth meta-review using GPT-4o and classify it into five sentiment categories: strongly positive, positive, neutral, negative, and strongly negative. To quantify emotional consistency, we use one-hot encoding to map the sentiment of each aspect to the vector space and concatenate these aspect-level vectors. Then, we calculate the cosine similarity between the generated emotional vector and the ground-truth emotional vector to obtain the emotional consistency score. In terms of content, we use ROUGE to measure text consistency (including ROUGE-1, ROUGE-2, ROUGE-L). ROUGE is a commonly used automatic evaluation metric primarily used to assess the quality of text in natural language generation tasks. It calculates similarity by comparing the overlap between the generated content and the true content. 
\section{Distribution of Score}
\subsection{Overall Distribution}
\begin{figure}[t]
\centering
\begin{subfigure}[!]{0.4\textwidth}
	%\caption{Restaurant}
	\includegraphics[width=\textwidth]{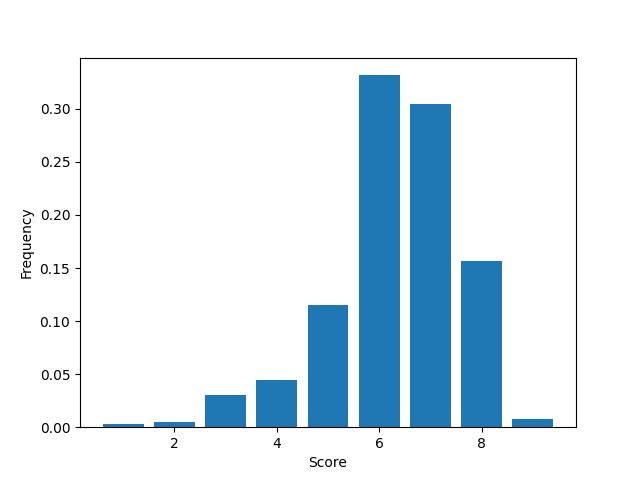}
\end{subfigure}
\begin{subfigure}[!]{0.4\textwidth}
%\caption{Restaurant}
\includegraphics[width=\textwidth]{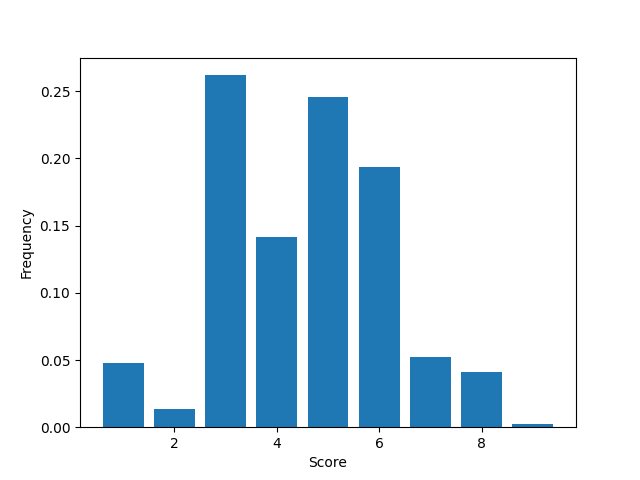}
\end{subfigure}
\begin{subfigure}[!]{0.4\textwidth}
	%\caption{Laptop}
	\includegraphics[width=\textwidth]{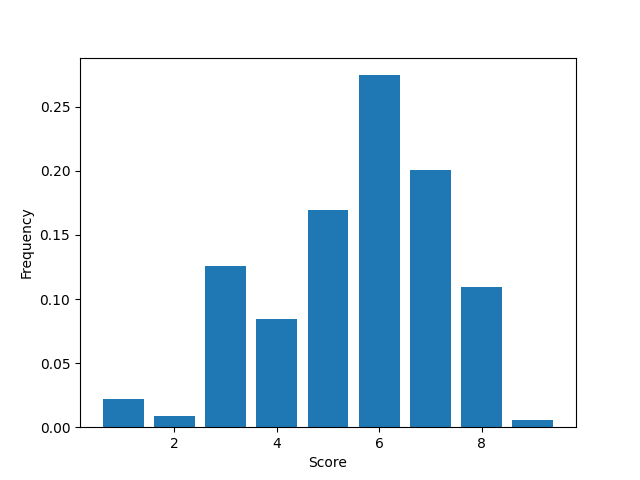}
\end{subfigure} 
\vspace{-10pt}
\caption{The distribution of score. }
\label{score_distribution}
\end{figure}

To preliminarily investigate the reviewer scores in the dataset, we analyzed the distribution of scores, as shown in the lower subfigure of Figure \ref{score_distribution}. The highest proportion of scores is 6, and the distribution follows a unimodal shape, with most scores concentrated between 3 and 8.
% \begin{figure}[t]
% \centering
% \includegraphics[width=\textwidth]{fig/score_overall.png}
% \vspace{-10pt}
% \caption{The distribution of score difference. }
% \label{para1}
% \end{figure}

\subsection{Conditional Distribution}
Based on the overall distribution, we also analyzed the conditional distribution of scores under the final decision of acceptance or rejection. Specifically, if a paper is accepted, all reviewer scores for that paper are recorded in this distribution. The upper subfigure of Figure \ref{score_distribution} illustrates the score distributions under the acceptance and rejection conditions. Under the acceptance condition, reviewer scores are more concentrated, with the vast majority of scores being 6 or higher, whereas the score distribution under the rejection condition is more dispersed, reflecting a greater divergence in reviewer opinions.

\section{Distribution of Score Difference}
\begin{figure}[t]
\centering
\begin{subfigure}[!]{0.4\textwidth}
%\caption{Laptop}
\includegraphics[width=\textwidth]{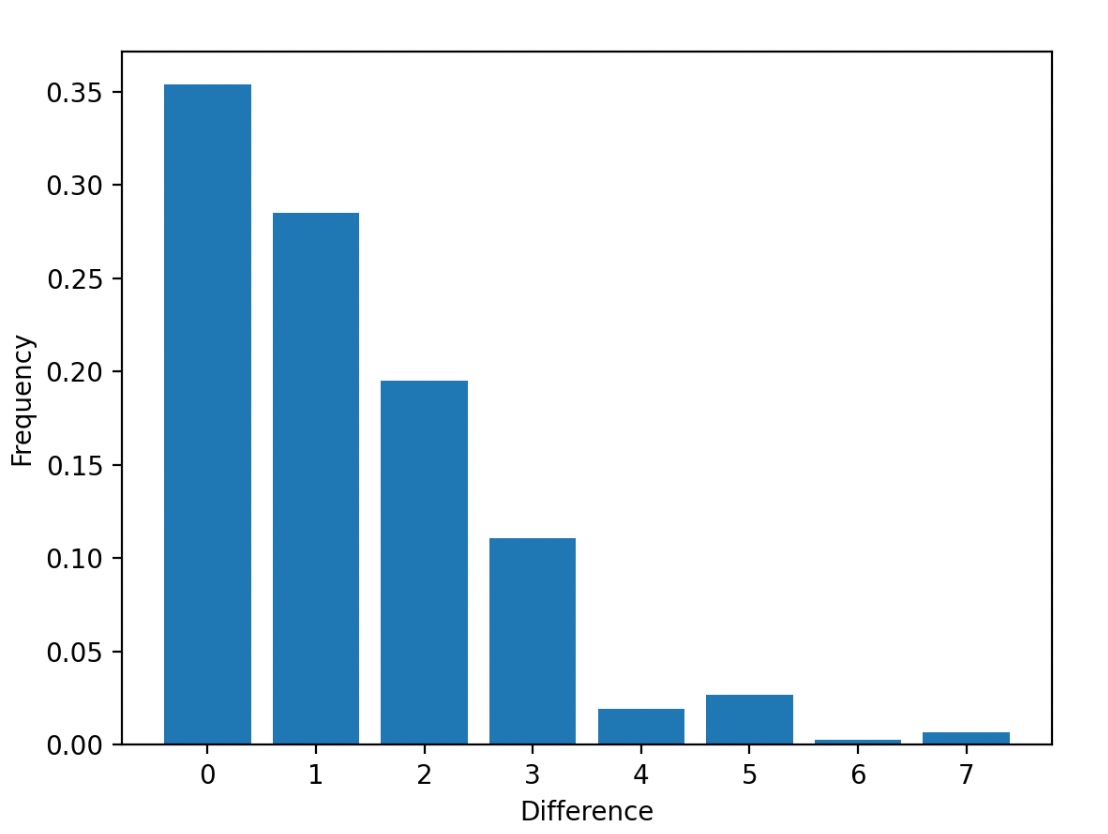}
\end{subfigure} 
\begin{subfigure}[!]{0.4\textwidth}
%\caption{Restaurant}
\includegraphics[width=\textwidth]{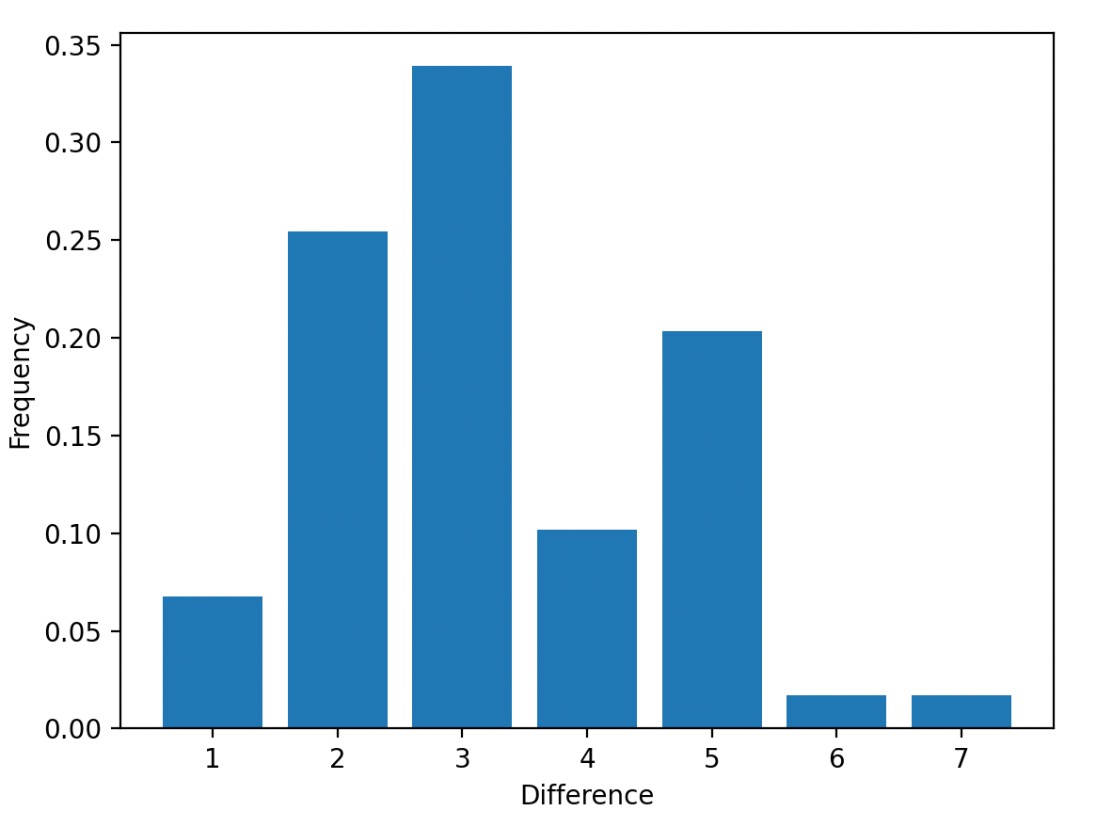}
\end{subfigure}
\caption{The distribution of score difference. }
\label{para1}
\end{figure}

From the frequency distribution of score differences across all data (the upper part of Figure \ref{para1}), it can be observed that the majority of reviewers assign scores with a difference of no more than 3 points for the same paper. This indicates that in real peer-review processes, reviewers generally exhibit a high level of agreement in their evaluations.\newline To investigate the conditions under which Rethink occurs, we introduce the conditional frequency distribution plot (the lower part of Figure \ref{para1}). The distribution of this plot significantly deviates from the overall frequency distribution, suggesting that the occurrence of Rethink is not independent of score differences. This finding indirectly supports the notion that the model can effectively identify conflicts among reviewers’ opinions and trigger Rethink at appropriate and necessary moments.

\section{Prompt Details}
\label{prompt}
\subsection{Aspects Definition}
To ensure objective and fair judgment criteria, we explicitly define the key evaluation dimensions: Novelty, Soundness, Clarity, Advancement, and Compliance, in Table \ref{tab4}. These definitions serve as a foundational framework for LLMs to accurately understand and assess the originality of the paper, its research logic, clarity of expression, contribution to the academic field, and adherence to academic standards. By providing clear and precise definitions, we aim to minimize ambiguities and subjectivity during the evaluation process. Furthermore, these definitions serve as essential guidelines, enabling LLMs to align their evaluation with the expectations of the academic community and to better support the peer-review process.

%In order to provide objective and fair judgment criteria, we explicitly present the definitions of Novelty, Soundness, Clarity, Advancement, and Compliance in Table \ref{tab4}. Clear definitions help LLMs accurately understand and evaluate the paper's originality, research logic, clarity of expression, contribution to the academic field, and compliance with academic standards.
\onecolumn
\begin{table*}[t]
    % \small
    \vspace{-20pt}
    \setlength\tabcolsep{3pt}
    \renewcommand{\arraystretch}{1}
    \centering
    % \begin{adjustbox}{max width=\linewidth}
    \begin{tabular}{p{2.5 cm}p{4.75cm}p{4.75cm}}
    \toprule
    \textbf{Aspect} & \textbf{Considerations} & \textbf{Definition} \\
    \midrule
     Novelty & 1.Model architectures \newline 2.New methodology \newline 3.Unexplored problem  & Assesses the originality and uniqueness of the research\\
    Soundness & 1.Mathematical correctness \newline 2.Reproducible experiment
 &Evaluates the theoretical and empirical validity of the proposed methods\\
    Clarity & 1.Well-structured writing \newline 2.Precise definitions \newline 3.Clear figures
 &Measures how well the paper conveys its ideas\\
    Advancement & 1.valid experiment  \newline 2.Ablation studies \newline 3.Compare with sota methods &Examines the extent to which the work pushes the field forward
\\
    % Sentiment Expresser & Original expresser of the sentiment &\\
    Compliance & 1.Proper citations \newline 2.Responsible AI \newline 3.Adherence to conference
  &Compliance ensures that the research adheres to ethical standards, dataset usage policies, and formatting guidelines.\\
    \bottomrule
    \end{tabular}
    % \end{adjustbox}
    \caption{Definitions of Aspects}
    \label{tab4}
\end{table*}

\subsection{Output Format}
To better parse the meta review data generated by the LLMs, we standardize the LLM's output into the following JSON format. In the following prompts, \texttt{\{json\_format\}} represents the JSON format.
\begin{lstlisting}[
breaklines=true, 
frame=single, 
numbers=left, 
numberstyle=\tiny,     % 行号字体更小
basicstyle=\ttfamily\footnotesize, % 代码字体更小
xleftmargin=2em,       % 增加左边距
linewidth=\textwidth,  % 限制代码块宽度
lineskip=-1pt,       % 减小行间距
columns=fullflexible,
] 
{
    ``aspects'':[
        {
        ``aspect'': ``Novelty'',
        ``opinion'': ``your opinion'',
        ``sentiment'': ``your sentiment'',
        ``convincingness'': ``your convincingness''
        },
        {
        ``aspect'': ``Soundness'',
        ``opinion'': ``your opinion'',
        ``sentiment'': ``your sentiment'',
        ``convincingness'': ``your convincingness''
        },
        {
        ``aspect'': ``Clarity'',
        ``opinion'': ``your opinion'',
        ``sentiment'': ``your sentiment'',
        ``convincingness'': ``your convincingness''
        },
        {
        ``aspect'': ``Advancement'',
        ``opinion'': ``your opinion'',
        ``sentiment'': ``your sentiment'',
        ``convincingness'': ``your convincingness''
        },
        {
        ``aspect'': ``Compliance'',
        ``opinion'': ``your opinion'',
        ``sentiment'': ``your sentiment'',
        ``convincingness'': ``your convincingness''
        },
        {
        ``aspect'': ``Overall quality'',
        ``opinion'': ``your opinion'',
        ``sentiment'': ``your sentiment'',
        ``convincingness'': ``your convincingness''
        }
    ]
    ``final_decision'': ``accept or reject''
}
\end{lstlisting}
\subsection{Sentiment Annotation}
\cite{li2024sentiment} has verified the high consistency between LLMs and humans in sentiment labeling. Therefore, this paper designs a prompt for automating the sentiment labeling of real meta reviews across various aspects and standardizing its format output.
\begin{lstlisting}[
breaklines=true, 
frame=single, 
numbers=left, 
numberstyle=\tiny,     % 行号字体更小
basicstyle=\ttfamily\footnotesize, % 代码字体更小
xleftmargin=2em,       % 增加左边距
linewidth=\textwidth,  % 限制代码块宽度
lineskip=-1pt,       % 减小行间距
columns=fullflexible,
] 
You are a professional meta-review content sentiment analyst. Please analyze the sentiments of all aspects of the meta-review: 
{meta_review}

Candidate criteria facets: Novelty, Soundness, Clarity, Advancement, Compliance, and Overall quality. 
Candidate sentiment levels: Strong negative, Negative, Positive and Strong positive
Candidate convincingness levels:  Not at all, Slightly Convincing, Highly Convincing
Format output according to the following JSON format.
{json_format}
\end{lstlisting}
\subsection{Prompt-CAF}
\subsubsection{Initial Prompt}
\label{Init_prompt}
In the initial stages of the model, critical points are extracted to streamline the synthesis process, minimizing information overload while preserving essential insights. These extracted elements are organized into a preliminary structure to support iterative refinement. The initial prompt used is provided below.
\begin{lstlisting}[
	breaklines=true, 
	frame=single, 
	numbers=left, 
	numberstyle=\tiny,     % 行号字体更小
	basicstyle=\ttfamily\footnotesize, % 代码字体更小
	xleftmargin=2em,       % 增加左边距
	linewidth=\textwidth,  % 限制代码块宽度
	lineskip=-1pt,       % 减小行间距
	columns=fullflexible,
	] 
You are an expert in reviewing scientific papers. Given the feedback provided by each reviewer, extract and summarize the key content in the following categories: (1) Strengths: List the positive aspects highlighted by the reviewer, such as novelty, methodology, or quality of results. (2) Weaknesses: List the areas considered problematic, such as unclear arguments, missing details, or flawed methodology.

\end{lstlisting}
\subsubsection{Prompt for First Review}
When the model reads the first review, it does not need to consider conflicts. Therefore, the prompt is designed as a simple viewpoint output format, as shown below.
\begin{lstlisting}[
breaklines=true, 
frame=single, 
numbers=left, 
numberstyle=\tiny,     % 行号字体更小
basicstyle=\ttfamily\footnotesize, % 代码字体更小
xleftmargin=2em,       % 增加左边距
linewidth=\textwidth,  % 限制代码块宽度
lineskip=-1pt,       % 减小行间距
columns=fullflexible,
] 
Assume you are a meta-reviewer. Below is the first reviewer's feedback on this paper. After reading the review, please generate your opinion on this paper.Output the response in JSON format.
{
    ``opinion'': ``your opinion'',
}
#########review########
review:{review_content}
score:{score}
confidence:{confidence}
\end{lstlisting}
\subsubsection{Prompt for Second Review}
When inputting a review that is not the first one, the model needs to consider whether the new review conflicts with its existing viewpoints and use this review to update its perspective. If a conflict exists, the viewpoints cannot be seamlessly integrated, and further consideration is needed on how to resolve the conflict. The prompt is designed as follows.
\begin{lstlisting}[
breaklines=true, 
frame=single, 
numbers=left, 
numberstyle=\tiny,     % 行号字体更小
basicstyle=\ttfamily\footnotesize, % 代码字体更小
xleftmargin=2em,       % 增加左边距
linewidth=\textwidth,  % 限制代码块宽度
lineskip=-1pt,       % 减小行间距
columns=fullflexible,
] 
Below is the {order} reviewer's feedback on this paper. After reading the review, please generate your opinion on this paper.After reading this review, please restate your opinion on the paper and indicate whether your current opinion conflicts with your previous one. Output the response in JSON format.
{
    ``opinion'': ``your opinion'',
    ``conflict'': ``yes or no''
}
#########review########
review:{review_content}
score:{score}
confidence:{confidence}
\end{lstlisting}
\subsubsection{Prompt for Cognitive Reconstruction}
We believe that conflicts arise from insufficient information, so we aim to resolve conflicts by adding key information. Compared to simple summarization, this “Slow Thinking” process provides the model with sufficient information, enabling it to better understand the relationships between the reviews and thus resolve the conflict. Therefore, we designed the following prompt starting from key concepts to address the conflict issue.
\begin{lstlisting}[
breaklines=true, 
frame=single, 
numbers=left, 
numberstyle=\tiny,     % 行号字体更小
basicstyle=\ttfamily\footnotesize, % 代码字体更小
xleftmargin=2em,       % 增加左边距
linewidth=\textwidth,  % 限制代码块宽度
lineskip=-1pt,       % 减小行间距
columns=fullflexible,
] 
Identify the relevant concepts causing the conflict between the opinions, and explain your understanding of these concepts.After this, restate your opinion and reconsider whether there are still any conflicts in the viewpoints.Format output according to the following JSON format.
{
    ``key_concepts'': ``your understanding of this concept'',
    ``opinion'': ``your opinion after recalling concept'',
    ``conflict'': ``yes or no''
}
\end{lstlisting}
\subsubsection{Prompt for Final Results}
After resolving all viewpoint conflicts, we use a simple prompt to have the model format and output its views on each aspect, along with the final opinion.
\begin{lstlisting}[
breaklines=true, 
frame=single, 
numbers=left, 
numberstyle=\tiny,     % 行号字体更小
basicstyle=\ttfamily\footnotesize, % 代码字体更小
xleftmargin=2em,       % 增加左边距
linewidth=\textwidth,  % 限制代码块宽度
lineskip=-1pt,       % 减小行间距
columns=fullflexible,
] 
You have resolved all conflicts between the reviews. Finally, pleasea aaa provide your final evaluation and acceptance decision based on the following six aspects: Novelty, Soundness, Clarity, Advancement, Compliance, and Overall Quality.Format output according to the following JSON format.
{json_format}
Candidate sentiment levels: Strong negative, Negative, Positive and Strong positive
Candidate convincingness levels:  Not at all, Slightly Convincing, Highly Convincing.As a strict and professional meta-reviewer, you should recommend acceptance if the paper meets the highest standards of academic excellence. The work should demonstrate significant novelty, provide rigorous theoretical or empirical contributions, and be methodologically sound without major flaws.Please output the target JSONL content directly without any additional content.
\end{lstlisting}
\subsection{Prompt-SCF}
This is a framework focused on improving the sentiment consistency between the model-generated meta review and the true values. It enhances multi-dimensional sentiment consistency through content extraction from various aspects and final summarization.
\begin{lstlisting}[
breaklines=true, 
frame=single, 
numbers=left, 
numberstyle=\tiny,     % 行号字体更小
basicstyle=\ttfamily\footnotesize, % 代码字体更小
xleftmargin=2em,       % 增加左边距
linewidth=\textwidth,  % 限制代码块宽度
lineskip=-1pt,       % 减小行间距
columns=fullflexible,
] 
Your task is to write a meta-review based on the following reviews and discussions for a scientific manuscript.

{input_documents}

Following the underlying steps below will get you better generated meta-reviews.

1. Extracting content and sentiment expressions of judgements in all above review and discussion documents;

2. Predicting Criteria Facets, Sentiment Levels, and Convincingness Levels;
Candidate criteria facets: Novelty, Soundness, Clarity, Advancement, Compliance, and Overall quality
Candidate sentiment levels: Strong negative, Negative, Positive and Strong positive
Candidate convincingness levels:  Not at all, Slightly Convincing, Highly Convincing

3. Reorganize extracted judgements in different clusters for different criteria facets;

4. Generate a small summary for judgements on the same criteria facet with comparison and aggregation;

5. Aggregate judgements in different criteria facets and write a meta-review based on the aggregation.

Format output according to the following JSON format.
{json_format}
\end{lstlisting}

\subsection{Prompt-Naive}
The naive prompt uses a simple command to instruct the model to output a formatted meta review.

\begin{lstlisting}[
breaklines=true, 
frame=single, 
numbers=left, 
numberstyle=\tiny,     % 行号字体更小
basicstyle=\ttfamily\footnotesize, % 代码字体更小
xleftmargin=2em,       % 增加左边距
linewidth=\textwidth,  % 限制代码块宽度
lineskip=-1pt,       % 减小行间距
columns=fullflexible,
] 
Your task is to write a meta-review based on the following reviews and discussions for a scientific manuscript.

{input_documents}

Format output according to the following JSON format.
{json_format}
\end{lstlisting}
\subsection{Prompt-Increm.}
\subsubsection{Intermediate Process Prompt}
The incremental naive prompt inputs the reviews to the model one by one, prompting the model to output a viewpoint after each input.
\begin{lstlisting}[
breaklines=true, 
frame=single, 
numbers=left, 
numberstyle=\tiny,     % 行号字体更小
basicstyle=\ttfamily\footnotesize, % 代码字体更小
xleftmargin=2em,       % 增加左边距
linewidth=\textwidth,  % 限制代码块宽度
lineskip=-1pt,       % 减小行间距
columns=fullflexible,
] 
Assume you are a meta-reviewer. Below is the {order} reviewer's feedback on this paper. After reading the review, please generate your opinion on this paper.Output the response in JSON format.
{
    ``opinion'': ``your opinion''
}
#########review########
review:{review_content}
score:{score}
confidence:{confidence}
\end{lstlisting}
\subsubsection{Increm.'s Generate Prompt}
After all the reviews are input, the final meta review is generated along with the received feedback.
\begin{lstlisting}[
breaklines=true, 
frame=single, 
numbers=left, 
numberstyle=\tiny,     % 行号字体更小
basicstyle=\ttfamily\footnotesize, % 代码字体更小
xleftmargin=2em,       % 增加左边距
linewidth=\textwidth,  % 限制代码块宽度
lineskip=-1pt,       % 减小行间距
columns=fullflexible,
] 
write a meta-review based on your previous opinion.

Format output according to the following JSON format.
{json_format}
\end{lstlisting}
\subsection{Prompt-SelfCoT}
\subsubsection{CoT Steps Generate Prompt}
The self-CoT method first uses the LLM to output the steps required to generate the meta review, and then has the LLM follow these steps to process all the input reviews and output the final feedback.
\begin{lstlisting}[
breaklines=true, 
frame=single, 
numbers=left, 
numberstyle=\tiny,     % 行号字体更小
basicstyle=\ttfamily\footnotesize, % 代码字体更小
xleftmargin=2em,       % 增加左边距
linewidth=\textwidth,  % 限制代码块宽度
lineskip=-1pt,       % 减小行间距
columns=fullflexible,
] 
{input_documents}

What are the steps to write a meta-review specifically for the above reviews and discussions of a manuscript.
\end{lstlisting}
\subsubsection{SelfCoT's Generate Prompt}
\begin{lstlisting}[
breaklines=true, 
frame=single, 
numbers=left, 
numberstyle=\tiny,     % 行号字体更小
basicstyle=\ttfamily\footnotesize, % 代码字体更小
xleftmargin=2em,       % 增加左边距
linewidth=\textwidth,  % 限制代码块宽度
lineskip=-1pt,       % 减小行间距
columns=fullflexible,
] 
{input_documents}

Follow the following steps and write a meta-review based on the above reviews and discussions for a manuscript.

{generated_steps}
Format output according to the following JSON format.
{json_format}
\end{lstlisting}
\newpage
\section{Case Study (Full Case)}
\label{case_study}
Below is the complete case from the case study in the main text, including the model's full thought process and the complete output.
\subsection{Our Method Context}
\begin{lstlisting}[
breaklines=true, 
frame=single, 
numbers=left, 
numberstyle=\tiny,     % 行号字体更小
basicstyle=\ttfamily\footnotesize, % 代码字体更小
xleftmargin=2em,       % 增加左边距
linewidth=\textwidth,  % 限制代码块宽度
lineskip=-1pt,       % 减小行间距
columns=fullflexible,
morekeywords={Step1, Step2,Step3,Final_output}, % 可以将特定关键字变成黑体
emphstyle=\bfseries\Huge        % 设置为黑体
]
Step1           
    __________________________________user_____________________________________
    
Assume you are a meta-reviewer. Below is the first reviewer's feedback on this paper. After reading the review, please generate your opinion on this paper.
review:
**(1) Strengths:**

- **Novelty**: The paper introduces a novel approach by using a color augmentation based on Planckian illuminants, which are physically realistic, rather than the existing standard color jitter methods.
- **Improved Performance**: The combination of networks trained with both standard color jitter and Planckian jitter shows significant improvements on classification tasks where color information can be beneficial.
- **Supporting Analysis**: The ablation study and analysis of results effectively substantiate the hypothesis that a physics-based color jitter improves the network's use of color information.
- **Relevance**: The paper emphasizes the importance of considering physics-based perspectives in data augmentation, addressing typical distributions of illumination in real-world data.
- **Applicability**: The method is seen as particularly useful in data sets where color plays a crucial role, such as flower, bird, and animal identification data sets.

**(2) Weaknesses:**

- **Lack of Direct Analysis**: The paper does not provide a direct analysis of the network's learned features or structure to verify if color features have indeed been utilized effectively.
- **Assumption of Relevance**: The assumption that color is a beneficial feature for the flower data set may not hold for other data sets, potentially limiting the generalizability of claims.
- **Limited Discussion**: There's insufficient discussion on alternative color distributions that could be better than Planckian illuminants and on other illumination scenarios.
- **Challenges with Material Color Augmentation**: The paper does not address the complexities associated with color augmentation for materials, which is class-based and not always appropriate for whole-image color changes.
- **Contextualization**: Including a discussion about the limitations of standard color augmentation methods could provide more context, although it is noted as not directly relevant to the paper's main contributions.
score:7
confidence:5
    ________________________________assistant___________________________________
    
opinion:The paper presents a novel approach to color data augmentation by introducing a physics-based method using Planckian illuminants, which appears to enhance the performance of deep networks on tasks where color information is crucial, such as object classification in datasets like flowers. The review highlights the importance of considering realistic illumination distributions in data augmentation, as opposed to uniformly randomized methods. The proposed method shows promise, particularly in scenarios where color plays a significant role, and the results are supported by an ablation study. However, the paper could benefit from a more direct analysis of the network's learned features to confirm the hypothesis. Additionally, while the focus on illumination color augmentation is well-justified, a discussion on the challenges of material color augmentation could provide valuable context. Overall, the paper makes a meaningful contribution to the field by addressing a limitation of standard color augmentation methods, and the reviewer's score of 7 reflects a positive assessment with room for further exploration and validation.

Step2
    __________________________________user_____________________________________
    
Below is the second reviewer's feedback on this paper. After reading this review, please restate your opinion on the paper and indicate whether your current opinion conflicts with your previous one.
review: 
(1) Strengths:
- The paper presents a novel physics-based approach to data color augmentation termed ``Planckian jitter'' which generates more color-realistic augmentations, especially for natural image classes such as birds and flowers.
- The method is well-demonstrated and motivated by following a standard black-body radiation model, which effectively constrains the color jitter to appear more realistic.
- Experiments are meticulously performed in a self-supervised manner, focusing on the impact of data augmentation on both natural and man-made datasets like Flowers-102, CUB-200, and CIFAR.
- Results depicted in figures and tables highlight the effectiveness of Planckian jitter in improving downstream classification accuracy, particularly for natural image classes at low resolutions.
- The paper extends the experiments to demonstrate that the proposed jitter approach is applicable to multiple self-supervised models.

(2) Weaknesses:
- The paper notes lesser effectiveness or even worse performance of Planckian jitter at higher resolution (224x244) for datasets such as Flowers-102 and VegFru, an issue that is not thoroughly explained or investigated in the paper.
- There is an ambiguous explanation related to color and shape discrimination at higher resolutions, suggesting a possible overlooked aspect of texture discrimination, which lacks detailed analysis.
- The results in Table 3 are somewhat counterintuitive compared to expectations from lower resolution results, and the paper does not provide a satisfying exploration of these discrepancies.
- While the paper discusses the combination of Planckian and random color jitter, it lacks a deeper investigation into making this combination more effective or class-specific, which could provide a more tailored approach to color augmentation.
- The paper acknowledges the overhead associated with combined color augmentation and mentions an incomplete investigation into the details of how the proposed method improves color realism or affects shape discrimination.
score:7
confidence:4
    ______________________________assistant_________________________________
    
opinion:The paper introduces a physics-based approach to color data augmentation, specifically using Planckian illuminants, which is shown to improve classification performance on natural image datasets like flowers and birds. The approach is well-motivated and demonstrated, with experiments conducted in a self-supervised setting across various datasets. The results indicate that Planckian jitter, especially when combined with random jitter, can enhance performance, although there are limitations, particularly at higher resolutions where the benefits are less clear. The paper is clear and direct in its contributions, and the experiments support the idea of more realistic color augmentation. However, the counter-intuitive results at higher resolutions and the need for more detailed analysis on the role of shape and texture at different resolutions suggest areas for further investigation. The suggestion to explore class-specific augmentation strategies could also enhance the approach's applicability. Overall, the paper makes a valuable contribution to the field, with a score of 7 reflecting its strengths and areas for improvement.
conflict:no

Step3
    __________________________________user_____________________________________
    
Below is the third reviewer's feedback on this paper. After reading this review, please restate your opinion on the paper and indicate whether your current opinion conflicts with your previous one.
review:
(1) Strengths:
- Utilizes realistic image processing, which is often overlooked in machine learning literature.
- The color model applied is largely accurate and well-grounded in color-conversion theory.

(2) Weaknesses:
- Questionable benefit of the Planckian jittering as demonstrated in the results, with concerns about the artificial manipulation of the testing dataset.
- Planckian jittering sometimes underperforms significantly as shown in the results.
- The motivation for using small image sizes (32x32) is unclear, and there's a need for testing on more realistic patch sizes.
- Misses crucial consideration of gamma encoding in the sRGB color space, potentially leading to inferior color correction results.
score:3
confidence:4
    ________________________________assistant___________________________________
    
opinion:The paper presents an innovative approach to color data augmentation using Planckian illuminants, aiming to improve self-supervised learning by providing more realistic color variations. While the method is grounded in color-conversion theory and shows potential benefits in tasks where color is crucial, such as flower classification, there are significant concerns about its effectiveness and implementation. The review highlights issues such as the artificial manipulation of test datasets, unclear motivations for using low-resolution images, and the lack of consideration for gamma correction in the sRGB space. These weaknesses raise questions about the robustness and general applicability of the proposed method. The mixed performance of Planckian jitter across different datasets and resolutions further suggests that the approach may not consistently outperform existing methods. Addressing these concerns, particularly the benefits of Planckian jitter and its impact on shape representation, is crucial for strengthening the paper's contributions. Overall, while the paper introduces a promising concept, it requires further validation and refinement to convincingly demonstrate its advantages.
conflict:yes
    __________________________________user_____________________________________
    
Identify the relevant concepts causing the conflict between the opinions, and explain your understanding of these concepts.After this, restate your opinion and reconsider whether there are still any conflicts in the viewpoints
    __________________________________assistant__________________________________
    
key_concepts:
The conflict arises from differing evaluations of the effectiveness and implementation of Planckian jitter for color data augmentation. Key concepts include: 
1) Planckian Jitter: A physics-based approach to color augmentation using Planckian illuminants, intended to provide more realistic color variations. While it shows potential in enhancing classification tasks where color is important, its performance is inconsistent across different datasets and resolutions. 
2) Test Dataset Manipulation: Concerns about the artificial manipulation of test datasets, which may not accurately reflect real-world conditions and could lead to misleading performance evaluations. 
3) Image Resolution: The use of low-resolution images (32x32) is questioned, as it may not provide sufficient information for practical applications. Higher resolutions could yield different results, as seen in the mixed performance at 224x224 resolution. 
4) Gamma Correction: The omission of gamma correction in the sRGB space is a critical oversight, as gamma compression affects color accuracy. Proper gamma correction is necessary for realistic color processing.
opinion:After considering these concepts, the paper introduces a promising approach to color data augmentation with Planckian jitter, but it requires further validation and refinement. Addressing concerns about test dataset manipulation, image resolution, and gamma correction is essential to demonstrate its advantages convincingly. The mixed performance across datasets and resolutions suggests that the method may not consistently outperform existing techniques. Overall, the paper contributes valuable insights but needs more robust evidence and analysis to support its claims.
conflict:no

Final_output
    __________________________________user_____________________________________
    
You have resolved all conflicts between the reviews. Finally, please provide your final evaluation and acceptance decision based on the following six aspects: Novelty, Soundness, Clarity, Advancement, Compliance, and Overall Quality.
    ________________________________assistant_______________________________
    
Novelty
opinion:The paper introduces a novel approach to color data augmentation using Planckian illuminants, which is a fresh perspective in the field.	
sentiment: Positive
convincingness: Highly Convincing
Soundness           
opinion:The methodology is grounded in color-conversion theory, but there are concerns about the robustness of the results due to issues like test dataset manipulation and lack of gamma correction.
sentiment: Negative,
convincingness: Slightly Convincing
Clarity
opinion:The paper is generally clear in its presentation, but some aspects, such as the motivation for low-resolution images and the impact of gamma correction, need more explanation.
sentiment:Positive
convincingness: Slightly Convincing
Advancement
opinion: The paper contributes to the advancement of data augmentation techniques, but the mixed performance across datasets and resolutions limits its impact.
sentiment:Negative,
convincingness:Slightly Convincing
Compliance
opinion:The paper complies with standard research practices, but the artificial manipulation of test datasets raises some ethical concerns.
sentiment:Negative
convincingness:Slightly Convincing
Overall quality
opinion:The paper presents an interesting idea with potential, but it requires further validation and refinement to meet the highest standards of academic excellence.
sentiment:Negative
convincingness: Slightly Convincing
Final_decision:reject
\end{lstlisting}
\newpage
\subsection{Naive Prompt Context}
\begin{lstlisting}[
breaklines=true, 
frame=single, 
numbers=left, 
numberstyle=\tiny,     % 行号字体更小
basicstyle=\ttfamily\footnotesize, % 代码字体更小
xleftmargin=2em,       % 增加左边距
linewidth=\textwidth,  % 限制代码块宽度
lineskip=-1pt,       % 减小行间距
columns=fullflexible,
] 
        __________________________________user_________________________________
    
Your task is to write a meta-review based on the following reviews and discussions for a scientific manuscript.

review1:The paper examines the color data augmentation used to train many deep networks for object classification tasks. It proposes that the existing color augmentation method does not permit the network to make effective use of color information when appropriate. They propose using a color augmentation that follows Planckian illuminants, which are physically-realistic. The paper shows that using the combination of a network trained using the standard color jitter and a network using the Plankcian jitter perform significantly better on classification tasks that can take advantage of color information. The ablation study and analysis of the results support the hypothesis that a physics-based color jitter enables the network to more effectively use color information. However, there is no direct analysis of the network's learned features/structure to confirm that. The author's assume that color is a useful feature on the flower data set--but is not necessarily a useful feature on other data sets--and use that assumption to make the claim. It is clear that the Planckian color jitter is responsible for the bump in performance on the flower data set, and the assumption is reasonable.
It's important to look at data augmentation from a physics-based perspective, because uniformly randomized data augmentation ignores the fact that illumination does have typical distributions. The existing data sets usually do not sample extensively from illumination distributions, especially given that most of the images are have white-balancing applied by default.
It would be nice to see more extensive use of physics-based principles, however, since almost all standard data sets use jpg-compressed and color enhanced images (designed for human viewing), most of the physical rules that govern illumination and materials are lost.  The Plackian jitter at least replaces the uniformly random color data augmentation with expected results on a data set where color is likely to play a strong role in classification. Other examples of data sets where that might be the case are the bird and animal identification data sets. Are there other color distributions, or ways of obtaining color distributions, that might be better than the Planckian illuminants?  There are illumination situations, such as light filtering through trees, that are non-Planckian but still non-zero in terms of likelihoods. There is a difference between color augmentation of materials and color augmentation of illumination. Illumination color is one color augmentation that can be done easily through a color rotation of the pixels. While on a jpg-compressed/color-enhanced image the effect is not perfect, it's close enough that the augmentation can allow the network to learn color-based rules that follow a more typical distribution of colors.  Regardless of whether color is a useful feature for an object, this type of augmentation matches what is likely to occur in the real world.\n\nColor augmentation for materials, on the other hand, is challenging to do properly as an augmentation. First, it is class-based, as every class will have a different material color distribution. Second, it is not appropriate to change the colors of the entire image in order to modify the material color. The uniformly random color augmentation seems like it is intended to try and cover both types of color changes (illumination and material).
It would be nice to have some discussion of this in the paper, though not necessary as it is not directly relevant to the author's contribution but an an additional limitation of the standard color augmentation methods.  But it would give a little more context to the issue in the introduction.
score:7
confidence:5

review2:This paper adds to the data augmentation transforms in the literature by proposing a more physics-based approach to generate more color-realistic augmentations compared to the standard more random color jitter.  This is well-demonstrated in Figure 1 for natural image classes such as flowers where although some classes may have strong color variations, in general the variations are more realistic when due to illumination, compared to man-made classes such as cars.  The approach is motivated by and follows a standard black-body radiation model, which again as shown in Figure 1 constrains the color jitter to be more realistic.\n\nThe experiments are performed in a self-supervised fashion as that nicely focuses the weight on data augmentation.  Both natural and man-made data sets are tested such as Flowers-102, CUB-200, and CIFAR.  
Overall the results summarized in Figure 3, and Tables 1, 2, 3 show the improvements in downstream classification with Planckian jitter on natural image classes and the potential for combining Planckian and random color jitter in general. The paper is very direct in the exact problem and contribution, namely more realistic data color augmentation in a self-supervised setting, especially for natural image classes such as birds and flowers.  For the targeted data sets such as Flowers-102, at low resolution 32x32, downstream classification can be improved by 5%.  Also by combining Planckian jitter with default random color jitter, similar or even larger improvements are shown more generally across data sets such as CIFAR and T1K+.  Overall the paper keeps it simple and clear for the reader and the overall idea is sound and the experiments very directly justify the basic idea of more realistic color augmentation for data sets with natural images.\n\nFigure 3 analyzing the color sensitivity of the results and the general results showing where for man-made imagery data sets, Planckian Jitter can be worse is useful to point out the limitations and details of the approach.  Finally it's an added bonus to verify that this basic general idea is applicable to multiple self-supervised models as shown in Table 4.  Although this isn't particularly surprising, it does add to the completeness of the experiments to support the basic idea in the paper.\n\nAlthough the LSC combined augmentation results show that Planckian and random jitter are somewhat complementary, the higher-resolution, 224x244, results in Table 3 do cause the reader to question whether the low resolution case is somehow special with respect to color augmentation.  Here Planckian jitter can actually be worse for the target data set of Flowers-102 which isn't fully explained why this happens.  
Although the idea seems ok for CUB-200 and better for T1K+, again for the natural image data set of VegFru, performance is worse or similar than default random color jitter.  The main explanation is a one sentence claim that for higher resolution shape is very representative and color may add little discrimination, but more investigation here is necessary.  For example, perhaps higher resolution is actually causing texture to be more discriminative, rather than shape?  Intuitively, shape seems reasonably preserved even at lower resolution, unless this is correlated with the actual self-supervised architecture of hyperparameters. As mentioned above, the main question for the reader is the somewhat counter-intuitive results in Table 3, for the same data sets but with higher resolution at 224x224.  One would expect a similar trend as the low-resolution data, but the numbers show otherwise.  Also the explanation on better shape discrimination on higher resolution bears more analysis and detail, e.g., is it more about texture than shape at higher resolution?\n\nAlthough the LSC combination is a good start to fuse both types of color augmentation, what about a way to make it more class-specific?  In other words, for even a coarse dichotomy such as natural vs man-made images, apply some complementary weighting/distribution of Planckian/default color jitter to get the best of both worlds? The paper has an explicit section on limitations regarding the overhead with the combined color augmentation approach and also mentions the incomplete investigation into how much and in what detail the proposed Planckian jitter improves color realism why discouraging shape discrimination.

review3:This paper proposes to use a realistic white balancing as a way to augment datasets for self-supervised learning approach. The balancing was based on Planckian blackbody radiation. The author used this to train a siamese model in conjunction with standard color jittering that involves hue and saturation perturbation, and showed that the combination of the two color jittering method works best. Furthermore, the planckian jittering seems to lead to better performance on tasks where color is an important features such as flower classification task. Strength:\n- Realistic image processing is complicated and is often skipped in ML literature. The color model used in this paper is largely correct and well-grounded in the color-conversion theory\n\nWeakness:\n- My main concern for this paper is that I am remain unconvinced of the benefit of the planckian jittering as presented:\n  - Fig 3 shows a more stable result, but it seems that the testing dataset also undergoes the simulated jittering rather than a real captures of different illumination. Such artificial manipulation of the test set is going to yield the same result regardless of how poorly the jittering performs.\n  - Planckian jitter do not always perform the best, and can underperform by a significant margin (see Table 1).\n  - The motivation for resizing images to 32x32 is unclear. In most practical application, such small patch sizes are unlikely to contain enough information to be useful. More investigation is needed on more realistic patch size of 200+pixel. In the present study, only Table 3 shows this result, and the performance of PJ is mixed on different dataset.\n- The color jittering model missed one crucial fact about the sRGB space, which is that it is gamma-encoded. Gamma compression is lossy, and resulted in far inferior color correction result compared to images captured in RAW. Nonetheless, I do not see the mentioning of gamma correction prior to color correction (and then reapplication to get back into the original color space). Please see my weakness section. I would like to see the response to my weakness section. particularly around the benefits of PJ, which is my main concern for this paper. Not quite. The author mentioned that their PJ reduces the quality of shape representation, but I do not see a convincing evidence of this in the text. No harm to the society is anticipated for the current manuscript.
score:3
confidence:4




    ________________________________assistant_______________________________

Novelty
opinion:The paper introduces a novel approach to color data augmentation by using Planckian illuminants, which is a more physics-based method compared to the standard random color jitter. This approach is innovative as it aligns with realistic illumination conditions.
sentiment:positive
convincingness:high
Soundness
opinion:The paper provides a sound theoretical basis for the proposed method, supported by experiments that demonstrate its effectiveness in certain scenarios. However, there are concerns about the lack of direct analysis of the network's learned features and the mixed results at higher resolutions.
sentiment:neutral
convincingness:medium
Clarity
opinion:The paper is generally clear in presenting its contributions and results. However, some aspects, such as the motivation for using low-resolution images and the impact of gamma correction, could be explained more thoroughly
sentiment:neutral
convincingness:medium
Advancement
opinion:The paper advances the field by proposing a more realistic color augmentation method that can improve classification tasks where color is a significant feature. It also highlights the potential for combining different augmentation methods.
sentiment:positive
convincingness:high
Compliance
opinion:The paper complies with standard scientific practices, and no ethical concerns are raised. However, it could benefit from addressing the limitations of the proposed method more comprehensively.
sentiment:positive
convincingness:medium
Overall quality
opinion:Overall, the paper presents a valuable contribution to the field of data augmentation with a novel approach. While there are some areas for improvement, particularly in the explanation of certain results and the handling of higher resolution images, the core idea is sound and well-supported by experiments.
sentiment:positive
convincingness:high
final_decision:accept
\end{lstlisting}
\subsection{The Ground Truth of Meta-Review }
\begin{lstlisting}[
breaklines=true, 
frame=single, 
numbers=left, 
numberstyle=\tiny,     % 行号字体更小
basicstyle=\ttfamily\footnotesize, % 代码字体更小
xleftmargin=2em,       % 增加左边距
linewidth=\textwidth,  % 限制代码块宽度
lineskip=-1pt,       % 减小行间距
columns=fullflexible,
] 
Training representations in computer vision typically requires systematic augmentations of the input training set such as crops, reflections, translations including color jitter. As illumination invariance tends to be a desired property in visual object detection tasks, design of specific color augmentations is of great interest. In this context, this paper proposes to improve the arguably inferior color representation of standard Color Jitter by proposing a realistic and convincing physics-based color representation schema, called Planckian Jitter and provides some experimental support. 
The discussion about the work concentrated around the overall effectiveness and experimental validation. The reviewers have mixed conclusions about the work. In light of the mixed opinions,I tend to agree with the concerns raised by reviewer W5kb, who suggests to 
-- Repeat experiments with higher resolution as 32 x 32 seems to be rather small for a conclusion of effectiveness.
-- Including some obvious baselines (e.g., no augmentation)
-- Clarify the take home message and conclusions from ablations.
I feel that the paper would benefit from a further iteration and that the current manuscript is not ready yet for publication.
\end{lstlisting}
			\end{document}